\def\year{2021}\relax
\documentclass[letterpaper]{article} % DO NOT CHANGE THIS
\usepackage{aaai21}  % DO NOT CHANGE THIS
\usepackage{times}  % DO NOT CHANGE THIS
\usepackage{helvet} % DO NOT CHANGE THIS
\usepackage{courier}  % DO NOT CHANGE THIS
\usepackage[hyphens]{url}  % DO NOT CHANGE THIS
\usepackage{graphicx} % DO NOT CHANGE THIS
\usepackage{times}
\usepackage{epsfig}
\usepackage{graphicx}
\usepackage{amsmath}
\usepackage{amssymb}

\usepackage{amsmath,amssymb} % define this before the line numbering.
\usepackage[normalem]{ulem}
\usepackage{subcaption}
\usepackage{algorithm}
\usepackage{algorithmicx}
\usepackage{algpseudocode}
\usepackage{multirow}
\usepackage{booktabs}
\usepackage[american]{babel}

\usepackage{natbib}  % DO NOT CHANGE THIS AND DO NOT ADD ANY OPTIONS TO IT
\usepackage{caption} % DO NOT CHANGE THIS AND DO NOT ADD ANY OPTIONS TO IT
\graphicspath{ {./graphics/} }

\title{DSAM: A Distance Shrinking with Angular Marginalizing Loss\\for High Performance Vehicle Re-identification}

\author{

   Jiangtao Kong, \textsuperscript{\rm 1}
   Yu Cheng, \textsuperscript{\rm 2}
   Benjia Zhou, \textsuperscript{\rm 3}
   Kai Li, \textsuperscript{\rm 1}
   Junliang Xing \textsuperscript{\rm 1}\\
}

\affiliations{
    \textsuperscript{\rm 1} Institute of Automation, Chinese Academy of Sciences \\
    \textsuperscript{\rm 2} National University of Singapore\\
    \textsuperscript{\rm 3} Macau University of Science and Technology \\
    tinysnowball0823@gmail.com, e0321276@u.nus.edu, 19098536ii20001@student.must.edu.mo, kai.li@ia.ac.cn, jlxing@nlpr.ia.ac.cn
}

\begin{document}

\maketitle

\begin{abstract}
Vehicle Re-identification (ReID) is an important yet challenging problem in computer vision. Compared to other visual objects like faces and persons, vehicles simultaneously exhibit much larger intraclass viewpoint variations and interclass visual similarities, making most exiting loss functions designed for face recognition and person ReID unsuitable for vehicle ReID. To obtain a high-performance vehicle ReID model, we present a novel Distance Shrinking with Angular Marginalizing (DSAM) loss function to perform hybrid learning in both the Original Feature Space (OFS) and the Feature Angular Space (FAS) using the local verification and the global identification information. Specifically, it shrinks the distance between samples of the same class locally in the Original Feature Space while keeps samples of different classes far away in the Feature Angular Space. The shrinking and marginalizing operations are performed during each iteration of the training process and are suitable for different SoftMax based loss functions. We evaluate the DSAM loss function on three large vehicle ReID datasets with detailed analyses and extensive comparisons with many competing vehicle ReID methods. Experimental results show that our DSAM loss enhances the SoftMax loss by a large margin on the PKU-VD1-Large dataset: $10.41\%$ for mAP, $5.29\%$ for cmc1, and $4.60\%$ for cmc5. Moreover, the mAP is increased by $9.34\%$ on the PKU-VehicleID dataset and $6.13\%$ on the VeRi-776 dataset. Source code will be released to facilitate further studies in this research direction.
\end{abstract}

\section{Introduction}
Vehicle Re-identification (ReID), which aims to match detected vehicle images across multiple non-overlapping cameras and different timestamps, has recently received great attention~\cite{ICCV17OIFEVehicleReID,yan2017exploiting,CVPR18ViewpointAwareVehicleReID,TMM18GroupVehicleReID,AAAI18VehicleReID}. It is an important computer vision task with many potential applications like intelligent traffic monitoring, intelligent video surveillance, and future smart city systems. It is also a very challenging problem due to factors like the large intraclass variations (e.g., in viewpoints, occlusions, illuminations, etc.), the invisibility of the license plate numbers (esp. when vehicles are of the same model and color), and the lack of large scale labeled training datasets (labeling identifications of vehicles is very difficult for human without using the license plate numbers).
\begin{figure}[t]
\begin{subfigure}{0.15\textwidth}
\centering
\includegraphics[width=0.92\linewidth]{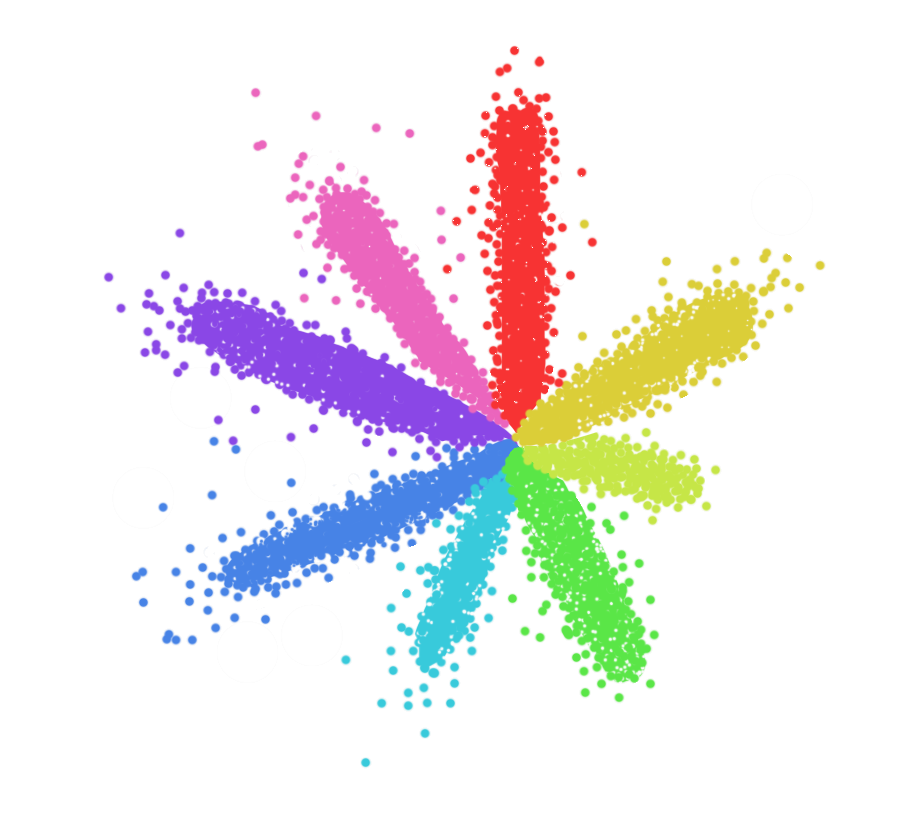}
\captionsetup{justification=centering}
\small
\caption{SoftMax in OFS}
\label{fig:softmax_eula}
\end{subfigure}
\begin{subfigure}{0.15\textwidth}
\centering
\includegraphics[width=0.9\linewidth]{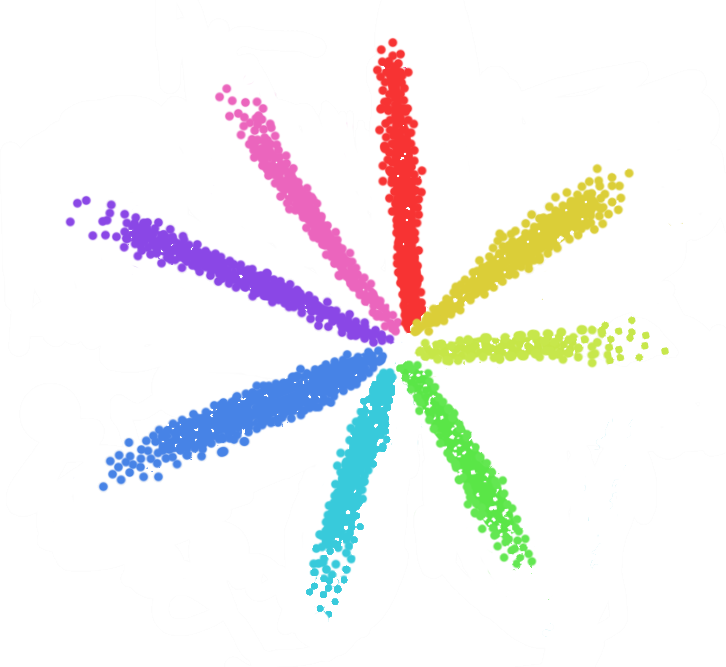}
\captionsetup{justification=centering}
\small
\caption{ArcFace in OFS}
\label{fig:arc_eula}
\end{subfigure}
\begin{subfigure}{0.15\textwidth}
\centering
\includegraphics[width=0.85\linewidth]{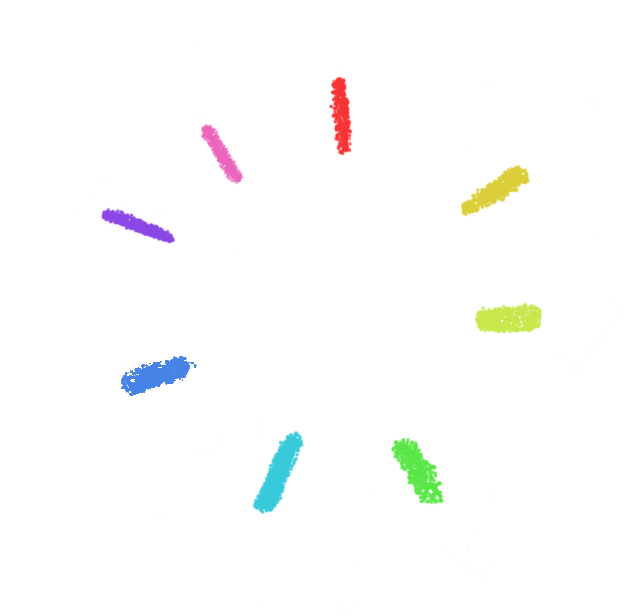}
\captionsetup{justification=centering}
\small
\caption{DSAM in OFS}
\label{fig:dsam_aeula}
\end{subfigure}
\begin{subfigure}{0.15\textwidth}
\centering
\includegraphics[width=0.83\linewidth]{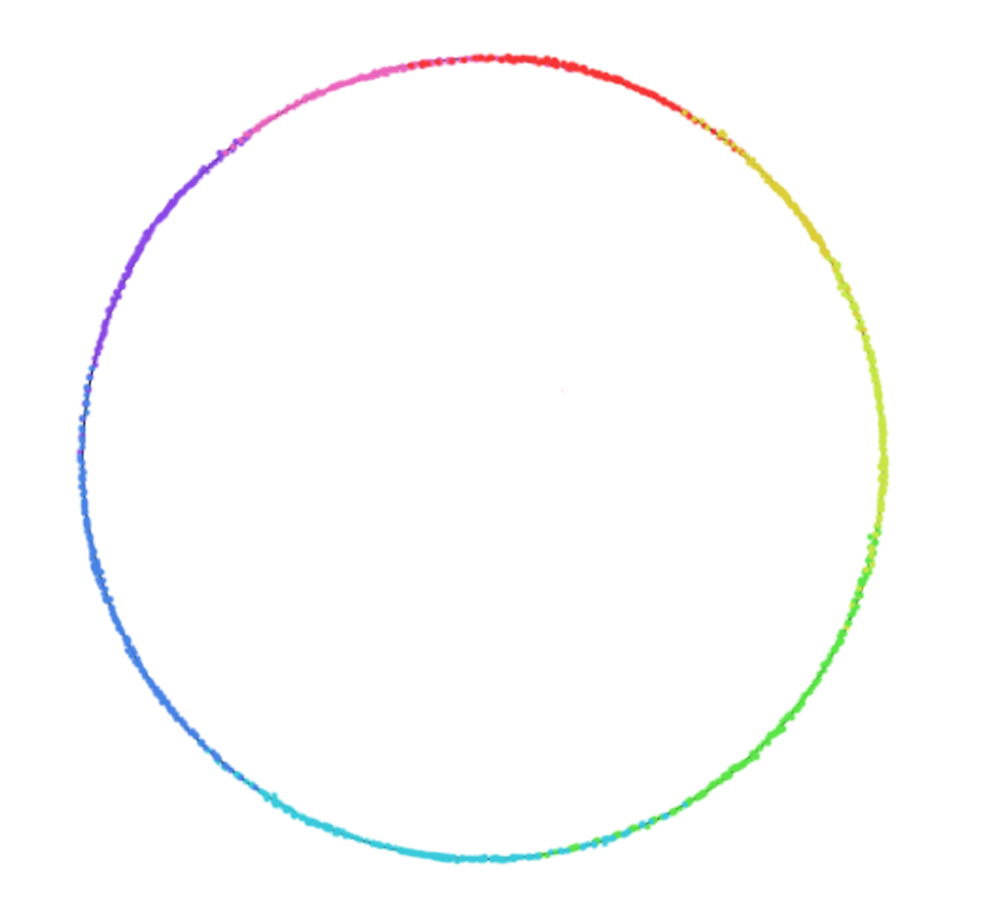}
\captionsetup{justification=centering}
\small
\caption{SoftMax in FAS}
\label{fig:softmax_angular}
\end{subfigure}
\begin{subfigure}{0.15\textwidth}
\centering
\includegraphics[width=0.9\linewidth]{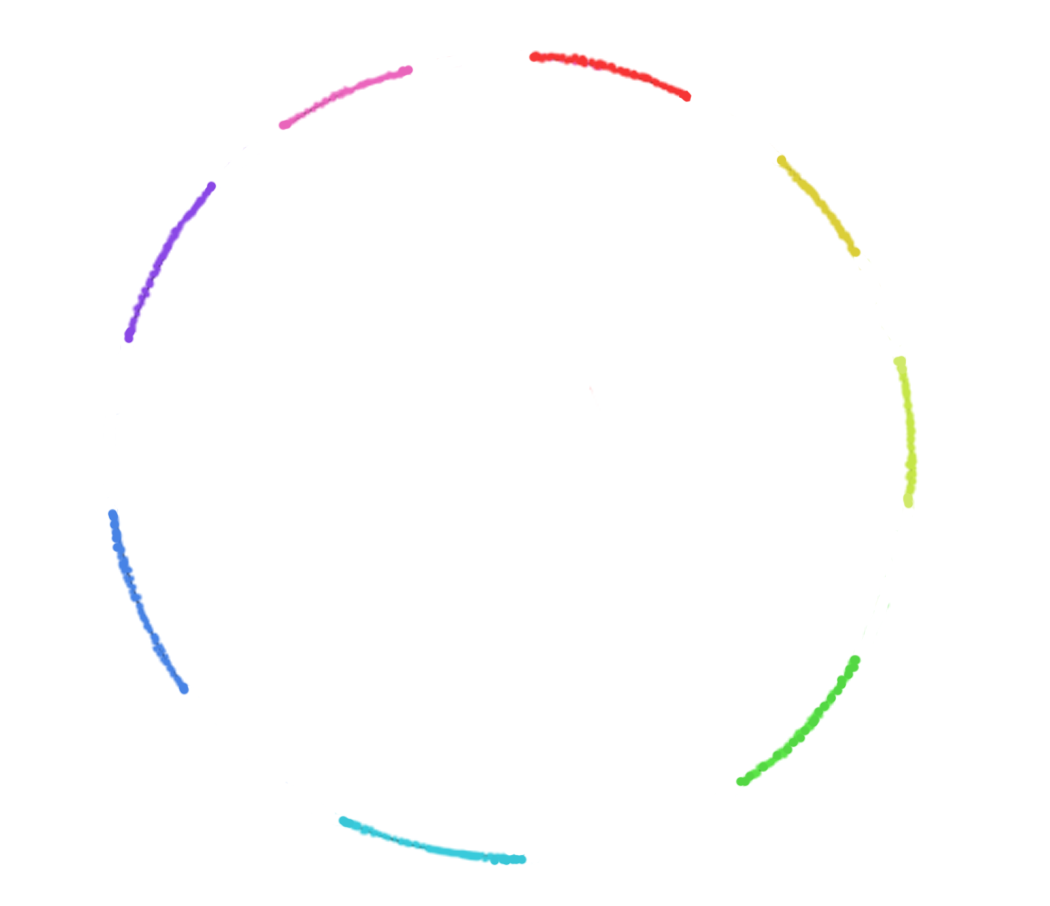}
\captionsetup{justification=centering}
\small
\caption{ArcFace in FAS}
\label{fig:arc_angular}
\end{subfigure}
\begin{subfigure}{0.15\textwidth}
\centering
\includegraphics[width=0.79\linewidth]{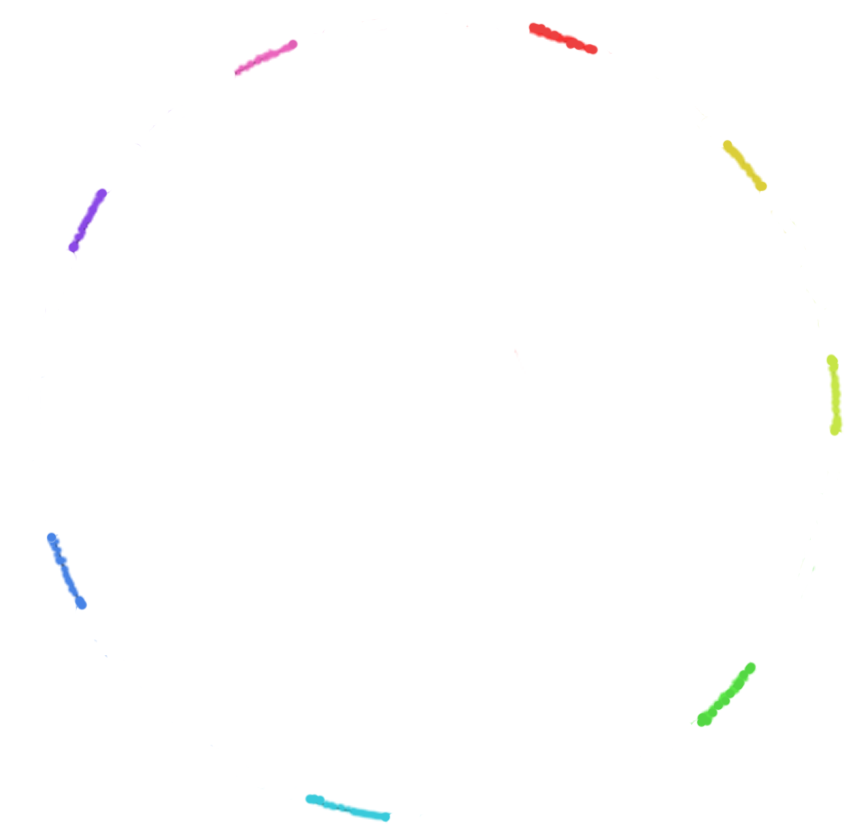}
\captionsetup{justification=centering}
\small
\caption{DSAM in FAS}
\label{fig:dsam_angular}
\end{subfigure}
%\captionsetup{font={footnotesize}}
\centering
\caption{A toy example to analyze the learning results of different loss functions. We select $8$ categories of vehicle images from the PKU-VehicleID dataset. Each category has about $6000$ images. As done in~\cite{liu2017sphereface,deng2019arcface,wang2018cosface}, we train different networks with different loss functions to embed the input images into 2D features. The features are visualized in both the OFS and the FAS.}
\label{fig:toy_example}
\end{figure}

With the fast development of deep neural networks, the performances of many visual recognition tasks have been greatly improved in the last decade, ranging from image classification~\cite{NIPS12AlexNet,CVPR16ResNet}, face recognition~\cite{CVPR15FaceNet,IJCV20FaceRecognition}, to person ReID~\cite{ICCV19SCAPReID,CVPR19ABDNet}. The dominant approach to recognizing the visual objects' identifications is adopting some neural network as the backbone model to learn the feature representations using different loss functions. The primary learning objective is to increase the interclass distance while decreasing the intraclass distance. Since there have been plenty of research efforts and achievements on the model architectures~\cite{NIPS12AlexNet,ICLR15VGGNet,CVPR16ResNet}, given a specified constraint on the backbone model, the main factor that accounts for the ReID performance is the design of the loss function.

The loss functions designed for visual recognition can be roughly classified into two groups. The first group is based on the multi-class classification formulation of the problem, which seeks to distinguish different classes' identifications as accurately as possible. Typical instances from this group include the original SoftMax loss, the ArcFace loss~\cite{deng2019arcface}, the CosFace loss~\cite{wang2018cosface}, and the SphereFace loss~\cite{liu2017sphereface}. The other group is based on the metric learning formulation of the problem, which seeks to maximize the interclass distance and intraclass similarity simultaneously. Typical examples in this group include the Triplet loss~\cite{CVPR15FaceNet} and the Center loss~\cite{wen2016discriminative}. The former group of loss functions are primarily designed for face recognition~\cite{NIPS14DeepID,deng2019arcface,wang2018cosface,liu2017sphereface}, where mainly the interclass distance is maximized. The latter group considers both interclass maximization and intraclass minimization, and has been widely used for face recognition~\cite{CVPR15FaceNet,wen2016discriminative} and person  ReID~\cite{Hermans2017In,Wojke2018deep}.

Compared to other visual objects such as faces and pedestrians, vehicles exhibit many distinct challenges to the identification task. The main reasons for these challenges come from the much larger intraclass diversities due to different acquisition viewpoints and the imperceptible interclass differences between the vehicles of the same model type and color.
This naturally raises a problem of whether the loss functions designed for face recognition or person ReID work well for the vehicle ReID task.
To answer this question, we train two models on an example dataset using the two most common loss functions in face recognition and person ReID, i.e., the original SoftMax loss and the ArcFace loss. As the results shown in Fig.~\ref{fig:toy_example}, the SoftMax does not work well in both the OFS and FAS, the ArcFace still has a large intraclass distance, which is detrimental to the performance of vehicle ReID. However, it does have a clear margin between different categories.
To further improve the loss function, the triple loss~\cite{CVPR15FaceNet} tries to increase the interclass distance and intraclass similarities for different combinations of the three training samples, two with the same identity, and the other one with a different identity. However, the triplet loss often suffers from slow convergence and poor local optima, partially because the triplet loss employs only one positive/negative example while not interacting with the other positive/negative examples per each update~\cite{sohn2016improved}.

To develop a more appropriate loss function to deal with the peculiar challenges in the vehicle ReID task, we in this work propose a Distance Shrinking with Angular Marginalizing (DSAM) loss function to improve the performance of deep vehicle ReID models with standard backbone networks. The proposed DSAM loss function performs hybrid learning in both the OFS and the FAS by incorporating the verification and identification signals into the training objective. By shrinking the vehicle training samples with the same identity per training batch online using the local information in the OFS and keeping a static margin between samples with different identities in the FAS from a global perspective, the samples with the same classes will be clustered into the pseudo-class centers automatically. The final distribution of feature representations preserves both a small intraclass distance and a clear boundary between different identities. Moreover, since the shrinking operation is performed in the batch level of the training samples during each training step of the model, it makes full use of local information so that can well suit for the distributions of the current feature representations.

Compared with existing loss functions for recognizing the object identities, the proposed DSAM loss function has the following main advantages:
\begin{itemize}
\item It maximizes the interclass distances and intraclass similarities from two feature spaces at the same time and leverages both the local and global information so that it well alleviates the specific challenges in the vehilce ReID problem.
\item It incorporates both the identification and verification signals into the learning objective with the iterated execution mechanism during each training step of the model, making it both adaptive and stable, and can speed up the training procedure notably.
\item It is straightforward to implement and does not introduce any additional learning parameters for the model, which can be integrated into all the existing SoftMax based loss functions to boost performance.
\end{itemize}
With the proposed loss function, we have presented a high-performance vehicle ReID model. Extensive experimental analyses and comparisons with many competing methods on three large vehicle ReID benchmarks demonstrate the effectiveness of the proposed loss function. On the PKU-VD1-Large~\cite{yan2017exploiting} dataset, DSAM loss enhances the performance of the SoftMax baseline by a large margin, i.e., $10.41\%$ for mAP, $5.29\%$ for cmc1, and $4.60\%$ for cmc5. On the PKU-VehicleID~\cite{CVPR16DRDLVehicleReID} and VeRi-776~\cite{liu2016large,ECCV16ProgessiveVehicleReID,liu2017provid} datasets, the mAP is enhanced by $9.34\%$ and $8.73\%$, respectively.
%In addition, we also do the experiment on the person ReID tasks, on the Market1501~\cite{Zheng_2015_ICCV} and DukeMTMC-reID~\cite{ristani2016performance}, the mAP is enhanced by $9.34\%$ and $8.73\%$, respectively.
To facilitate further studies in the vehicle ReID problem, the source code, trained models, and all the experimental results will be released.

\section{Related Work}
The visual object ReID task aims to find the target object in a large scale gallery with significant progress made in the last decade. Many hand-crafted features are proposed to capture the visual features~\cite{Chen2016Similarity,Liao2015Person,shi2015transferring} and CNN-based features~\cite{Cheng_2016_CVPR,Zhao_2017_CVPR}, significantly improving the person ReID performance. In the following, we mainly discuss the work related to vehicle ReID.

\textbf{Vehicle ReID.} Under unlimited conditions, public large-scale vehicle ReID datasets~\cite{liu2016large,AAAI18VehicleReID,lou2019veri,ECCV16ProgessiveVehicleReID,CVPR16DRDLVehicleReID} with annotation labels and rich attributes are widely collected. These datasets present different challenges in terms of occlusion, lighting, low resolution, and various views. One way to deal with these challenges is to use deep features~\cite{ICCV17OIFEVehicleReID,ECCV16ProgessiveVehicleReID} instead of hand-crafted features to describe vehicle images. To learn more powerful features, some methods~\cite{AAAI18VehicleReID,ECCV16ProgessiveVehicleReID,liu2017provid,wei2018coarse,shen2017learning} try to explore the details of the vehicle using additional attributes (such as model type, color, space-time information, etc.). Besides, adversarial learning-based methods~\cite{lou2019embedding,zhou2017cross} use a synthetic multi-view vehicle image adversarial network to mitigate the cross-field effect between vehicles. Multi-view learning-based methods~\cite{CVPR18ViewpointAwareVehicleReID,ICCV17OIFEVehicleReID} implement view-invariant inference by learning the representation of perceptual points.

In addition to learning global vehicle features, a series of part-based methods explicitly utilize discriminative information from multiple vehicle parts. Region-based methods~\cite{liu2018ram,zhu2019vehicle,chen2019multi} divide the feature map into multiple partitions to extract local feature representations of each region. Attention based~\cite{khorramshahi2019dual,kanaci2019multi,khorramshahi2019attention} uses key points information to pay more attention to more effective local features.

\textbf{Loss Functions.} In addition to the network structures, the loss functions also play an important role in learning effective feature representations for the ReID task. Different kinds of loss functions have been proposed in the last decades for face recognition and person ReID~\cite{wang2018support,Chen_2017_CVPR,zhu2020aware}. Among them, the most widely used loss function is the SoftMax loss~\cite{NIPS12AlexNet}, which employ the class identification signal from the global perceptive. Following improvements over the SoftMax loss include CosFace~\cite{wang2018cosface}, ArcFace~\cite{deng2019arcface}, and SphereFace~\cite{liu2017sphereface}, which enhance the global identification signals by adding different constraints in some angular spaces of the feature representations. These loss functions obtain quite good classification results of the object identities. However, it ignores the objects' local intraclass variations, which impairs the ReID performance, especially for objects with large intraclass variations. The Contrastive loss~\cite{NIPS14DeepID} and Triplet loss~\cite{CVPR15FaceNet} use the verification signal locally to increase the Euclidean margin for better feature embedding. The Center loss ~\cite{wen2016discriminative} learns the feature representation center for each identity and uses these centers to reduce the intraclass variations. All the above-discussed loss functions provide excellent insights from a specific perspective for designing effective loss functions. However, they all ignore some factors which will be useful to enhance the loss function further.

In Table~\ref{table:losses}, we summarize these loss functions by characterizing them from different perspectives, including the use of local and/or global information, the feature space used, the optimization of the interclass and/or intraclass variations. From Table~\ref{table:losses}, we can observe that the proposed DSAM loss function with SoftMax provides a comprehensive solution with all the factors considered, which optimize the intraclass and interclass variations simultaneously from both the OFS and FAS using the local information and global information at the same time.

\begin{table}
\small
%\captionsetup{font={footnotesize}}
\centering
\begin{tabular}{cccc}
%\hline
\toprule
Loss &Local/Global &OFS/FAS &Inter/Intra \\ \hline
S &Global &FAS &Inter \\
AF/CF/SF &Global &FAS &Inter \\
T &Local &OFS &Both \\
S-based+T &Both &OFS &Both \\
C &Local &OFS &Inter \\
S-based+C &Both &OFS &Both \\ \hline
DSAM &Local &Both &Both \\
S-based+DSAM &\textbf{Both} &\textbf{Both} &\textbf{Both} \\
%\hline
\bottomrule
\end{tabular}
\caption{Comparison of different loss functions. The ``S" means SoftMax, ``AF", ``CF", and ``SF" denote ArcFace, CosFace, and SphereFace respectively. The ``C" means center loss and the ``T" means triplet loss, DSAM is the proposed loss function.}
\label{table:losses}
\end{table}

\section{Proposed Approach}
\subsection{Motivation}
\label{section3.1}
We started by analyzing different SoftMax based loss functions. The original SoftMax loss is calculated as follows:
\begin{equation}
\label{formual_SoftMax}
L_\text{SoftMax}=-\frac{1}{N}\sum_{i=1}^N \log \frac{e^{f_{y_i}}}{\sum_{j=1}^n e^{f_j}},
\end{equation}
where $N$ denotes the batch size, ${y_i} \in [1,n]$ is the label of the $i$-th sample, $n$ denotes the class number, $f_j$ and $f_{y_i}$ are the scores of the $j$-th class and the $y_i$-th class.
The performance of the SoftMax loss function on ReID tasks is not so good since it can not deal with the hard samples well.
For example, suppose that the scores $f_j$ are the same for $j\in[1,n], j\neq y_i$, and the score of the $y_i$-th class $f_{y_i}$ is $z$ ($z\ge1$) times of $f_j$, then the probability $P_{y_i}$ of the $i$-th sample belong to the $y_i$-th class is calculated as:
\begin{equation}
\label{formual_scores}
P_{y_i}\!=\!\frac{e^{f_{y_i}}}{\sum_{j=1}^n e^{f_j}}\!=\!\frac{e^{zf_{j}}}{\sum_{j=1}^n e^{f_j}}\!=\!\frac{1}{1+(n-1)e^{(1-z)f_{j}}}
\end{equation}
The value of $P_{y_i}$ tends to approach $1$ with large $f_j$, even for hard samples that have relatively small $z$. In Eq.~(\ref{formual_derivative}), $W_{y_i}$ denotes the $y_i$-th class's weight vector, since the value of $P_{y_i}$ is very close to $1$ for the hard samples, the gradient $\frac{\partial L_{\text{SoftMax}}}{\partial W_{y_i}}$ will vanish:
\begin{equation}
\label{formual_derivative}
\frac{\partial L_{\text{SoftMax}}}{\partial W_{y_i}}=\big(P_{y_i}-1\big)x.
\end{equation}

Based on the above analyses, the model trained by SoftMax loss is easy to get saturated, which results in ambiguous boundaries between different classes since the hard samples near boundaries can contribute little to the training process. In order to obtain clear margins between different categories, the ShpereFace~\cite{liu2017sphereface}, CosFace~\cite{wang2018cosface} and ArcFace~\cite{deng2019arcface} change the original form of the SoftMax loss in an angular margin manner. In ReID tasks, the cosine distance between the probe-gallery pair is used for testing. Thus the angular margin loss, which defines a clear margin in the FAS, often works well:
\begin{equation}
    \centering
    \label{fyi_ArcFace}
    f_{y_i}=s*\cos({m_1\theta_{y_i}}+m_2)+m_3,
\end{equation}
\begin{equation}
    \centering
    \label{formula_ArcFace}
    L_\text{ang-margin}=-\frac{1}{N}\sum_{i=1}^N \log \frac{e^{f_{y_i}}}{e^{f_{y_i}}+\sum_{j=1,j\neq y_i}^n e^{s*\cos{\theta_j}}},
\end{equation}
where $s$ denotes the hypersphere scale, and $m$ denotes the angular margin penalty between the feature vector $x_i$ and the class weight vector $W_i$. As Eq.~(\ref{fyi_ArcFace}), Eq.~(\ref{formula_ArcFace}) shows, the angular margin loss normalize $W_i$ and $x_i$ by $\ell_2$ norm and remove the bias term, so that the loss function can focus on $\theta_{y_i}$ and $\theta_{j}$ and use the margin penalty $m$ to enlarge the distance between different classes.

However, the angular margin loss is not producing optimal results. The samples within the $y_i$-th class's boundary will have a much larger $f_{y_i}$ than the other classes' score $f_j$ since the $s$ value is always set to a relatively large value for better convergence~\cite{wang2017normface}, the samples located in the marginalized boundary is easy to saturate in terms of $P_{y_i}$. As a result, the gradient will vanish, and no further update will be performed to enhance the intraclass compactness. In order to obtain a more compact distribution with a smaller intraclass distance than the angular margin loss does, we propose our DSAM loss in the next section.

\subsection{The Proposed DSAM Loss}
\label{section3.2}
\begin{figure}[t]
\centering
\begin{subfigure}{0.23\textwidth}
\includegraphics[width=\linewidth]{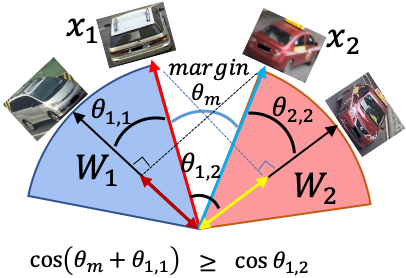}
\caption{ArcFace}
\label{fig:ArcFace_compare}
\end{subfigure}
\begin{subfigure}{0.23\textwidth}
\includegraphics[width=\linewidth]{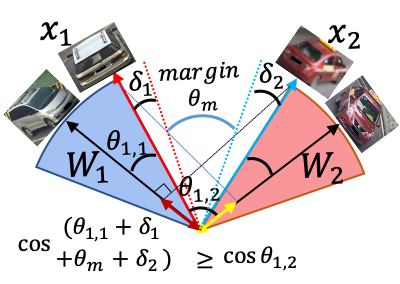}
\caption{DSAM}
\label{fig:DSAM_compare}
\end{subfigure}
%\captionsetup{font={footnotesize}}
\caption{A comparison of different loss functions' decision boundaries. The colored areas represent the decision boundary of different classes, and the DSAM gets the largest margin between classes in two loss functions by narrowing the intraclass distance.}
\label{fig:compare}
\end{figure}

As deep embedding features always distribute around the weight $W_i$ which can represent the center of features belong to the same ID in the hypersphere~\cite{deng2019arcface}, we can shrink the angle $\theta_{W_i, x_i}$ between the weight $W_i$ and the feature $x_i$ to narrow the intraclass distance in angle space.
%\begin{figure}[t]
%\captionsetup{font={footnotesize}}
%\centering
%\includegraphics[width=0.47\textwidth]{theory.png}
%\caption{The yellow points belong to the same ID and the green points which denote the negative samples are belong to another ID. The left one shows the process of shrinking the intraclass distance in the OFS and the middle one shows how to keep a margin between classes in the FAS, the last one show the final results of this process.}
%\label{fig:theory}
%\end{figure}
In order to carry out the metric learning scheme, in each batch of training samples, we draw $P$ classes and $k$-th class's samples form a set $\mathbb{Q}_{k}$ with the number of samples of $Q$, and we define $\mathbb{Q}_{all}=\{1\ldots P*Q\}$ is a set include all samples in per batch.
\begin{equation}
\mathbb{A}_p^a\!=\!\{z\in\mathbb{Q}_{all} | y_z=y_a\},\mathbb{A}_n^a=\{z\in\mathbb{Q}_{all} | y_z\neq y_a\},
\end{equation}
\begin{equation}
L^a_{pos} =\sqrt{\sum_{i\in \mathbb{A}_p^a}\|X_a-X_i\|^2},
\label{eq:posloss}
\end{equation}
where $\mathbb{A}_p^a$ and $\mathbb{A}_n^a$ are sets of $X_a$ 's positive samples and negative samples in per batch, $y_z=y_a$ denotes that $X_z$ and $X_a$ belong to the same class and $X_a$ is the anchor feature, and $X_i$ are features with the same label as an anchor in each batch. This loss term aims to minimize the distance between positive samples in the OFS. When the Euclidean distance between the positive samples decreased, samples like $x_2$, $x_3$, $x_4$ move towards to $x_1$ which makes $\theta_{W_i,x_1}$ approach to $0$. In summary, by minimizing the $L^a_{pos}$, the $\theta$ between samples and the $\theta_{W_i,x_i}$ can be minimized simultaneously, which will result in the small intraclass distance in both the original and angular space as shown in Fig.~\ref{fig:DSAM_compare}.
\begin{figure}[t]
%\captionsetup{font={footnotesize}}
\centering
\includegraphics[width=0.47\textwidth]{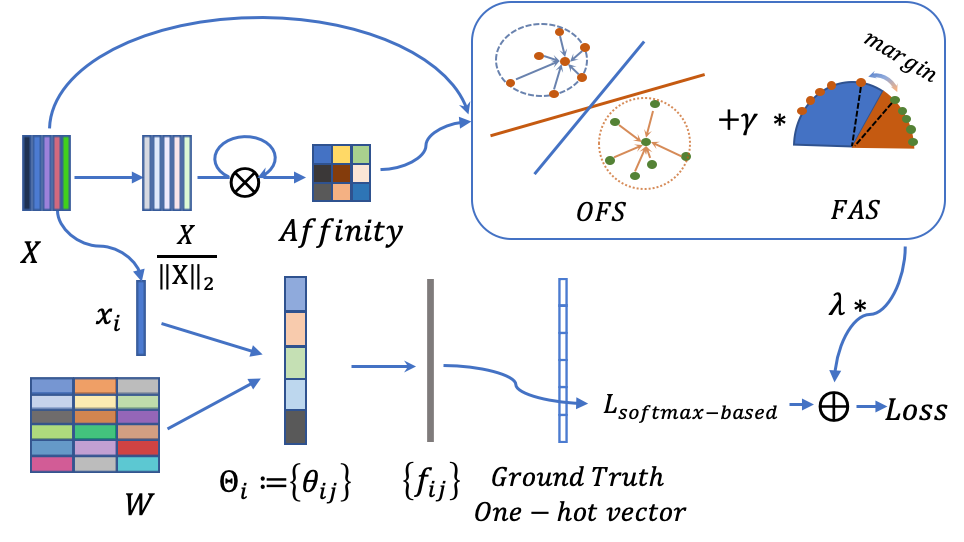}
\caption{The whole pipeline of training a DCNN model for vehicle ReID supervised by our DSAM loss. Both the original and the normalized feature vectors are used to compute the DSAM loss. We exploit all positive samples to compact the intraclass distance and use the furthest positive sample to keep a margin between negative samples as Eq.~(\ref{DSAM_loss}) does. Finally, we weighted combine the DSAM loss with the SoftMax based loss as the final loss function.}
\label{fig:process}
\end{figure}

To keep clear boundaries between different classes in the FAS, we deal with it in the FAS directly. We use $D_{i,j}$ to denote the difference between $X_i$ and $X_j$ in FAS as:
\begin{equation}
    \centering
    \label{formula_diff}
    D_{i,j}=e^{2-2\frac{{X_i}^T}{\|X_i\|_2}\frac{X_j}{\|X_j\|_2}}-1.
\end{equation}

The $D_{i,j}$ is a monotonically increasing function with respect to the angular distance between sample $X_i$ and $X_j$. Since the hardest sample most likely represents the lower bound of the distance between the negative class and positive class, we use $\max\limits_{j\in\mathbb{A}_p^a}D_{a,j}$ to choose the farthest positive sample for an anchor in each batch and add an additive margin penalty, which is formulated as
\begin{equation}
    L^a_{neg}\!=\!\frac{1}{(P-1)Q} \!\sum_{i\in\mathbb{A}_n^a}\max\big(0,m_{neg}-(D_{a,i}-\max\limits_{j\in\mathbb{A}_p^a}D_{a,j})\big),
    \label{neg_loss}
\end{equation}
where $i\in\mathbb{A}_n^a$ means we only focus on negative samples with different ground-truth labels from the anchor. The loss for negative samples aims to enlarge the angular distance to the pre-defined margin $m_{neg}$. By doing so, clear boundaries between different classes can be obtained.

Generally, our DSAM loss is the combination of positive and negative loss terms with a constant $\gamma$ and is formulated as:
\begin{equation}
    L_\text{DSAM} = \frac{1}{PQ} \sum_{k=1}^P\sum_{a\in \mathbb{Q}_k} (L^a_{pos} + \gamma  L^a_{neg}).
    \label{DSAM_loss}
\end{equation}

In practice, our DSAM loss is used with SoftMax based loss for model training. The SoftMax based loss can find class centers for each class using global information, and our DSAM loss can cluster the samples to their corresponding centers for a more compact feature distribution. The final loss is defined in Eq.~(\ref{final_loss}) and $\lambda$ is the weight for DSAM. Fig.~\ref{fig:process} shows the whole training process of our DSAM:
\begin{equation}
    \centering
    \label{final_loss}
    L=L_\text{softmax-based}+\lambda L_\text{DSAM}.
\end{equation}

\subsection{Discussion}
\label{section3.3}
In this subsection, we firstly discuss the reason why we use Euclidean distance but not angular distance in $L^a_{pos}$, and then illustrate how our DSAM impacts the decision boundary. Finally, we also visualize the comparison between the effect of DSAM loss and the effect of the angular margin loss functions.

In order to shrink the angle $\theta_{W_i, x_i}$ between the weight $W_i$ and the feature $x_i$ to narrow the intraclass distance in angle space, it is very natural to use the angular distance directly to decrease the intraclass distance. Assuming that the loss called $L^a_{ang-pos}$ uses the angular distance directly and $L^a_{ang-pos}$ is defined as follows,
\begin{equation}
    \cos{\theta_a^i}:=\frac{{X_a}^T}{||X_a||_2}\frac{X_i}{||X_i||_2}=x_a\cdot x_i,
    \label{eq:costheta}
\end{equation}
\begin{equation}
    L^a_{ang-pos}=\sum_{i\in\mathbb{A}_p^a}\arccos({x_a\cdot x_i}),
    \label{eq:loss_ang_pos}
\end{equation}
where $x_a$ and $x_i$ means the $X_a$ and $X_i$'s unit vectors which get from Eq.~(\ref{eq:costheta}), decreases the intraclass distance by minimizing the angular distance directly and the gradient can be calculated as:
\begin{equation}
    \frac{\partial L^a_{ang-pos}}{\partial x_i}=-\frac{x_a}{\sqrt{1-(\cos{\theta_a^i})^2}}.
    \label{eq:loss_ang_gradient}
\end{equation}
In many large-scale datasets, angles between positive samples are very small, and the $\cos{\theta_a^i}$ is limited to $1$. $L^a_{ang-pos}$ is hard to optimize because of the exploded gradient when $X_a$ and $X_i$ are very close, so we use $L^a_{pos}$ to minimize the distance between positive samples in the OFS.

In the FAS, as shown in Fig.~\ref{fig:compare}, we use the $\ell_2$-normalized class weight vector $W_1$ and $W_2$ to represent the center of the class $C_1$ and $C_2$. $x_1$ and $x_2$ denote the $\ell_2$-normalized features in different classes, and the angle between $x_i$ and $W_j$ is $\theta_{i,j}, i\in\{1,2\}, j\in\{1,2\}$.

The Softmax loss results in a side-by-side decision boundary and hard samples near the boundary are hard to be separated. Similar effect is also observed in Fig.~\ref{fig:toy_example}. As shown in Fig.~\ref{fig:compare}, by adding margins in the angular perspective, which is defined in the angular margin losses as $\cos(m_1 \theta_{1,1} + m_2) + m_3 = \cos(\theta_{1,1} + \theta_{m}) \ge \cos \theta_{1,2}$, we can obtain a clearer decision boundary where samples are well separated, but the intraclass variance is still large due to gradient vanish discussed in the previous section. As for our DSAM loss, a larger margin between class $C_1$ and $C_2$ can be obtained by reducing the intraclass variance. Assume the intraclass variance is reduced by $\delta_{i}$ for each class, the decision boundary produced by our DSAM loss will be:
\begin{equation*}
\cos(\theta_{1,1} + \arccos(m_{neg}) + \delta_1 + \delta_2) \ge \cos(\theta_{1,1} + \theta_{m}),
\end{equation*}
where $\theta_{m}$ is the margin angle produced by angular margin loss which is equivalent to negative sample margin $\arccos(m_{neg})$ in our DSAM. Based on the result, it is clear that our DSAM loss obtains a lager margin than the angular margin loss by shrinking the intraclass distance.
\begin{figure}[t]
\centering
\begin{subfigure}{0.23\textwidth}
\includegraphics[width=\linewidth]{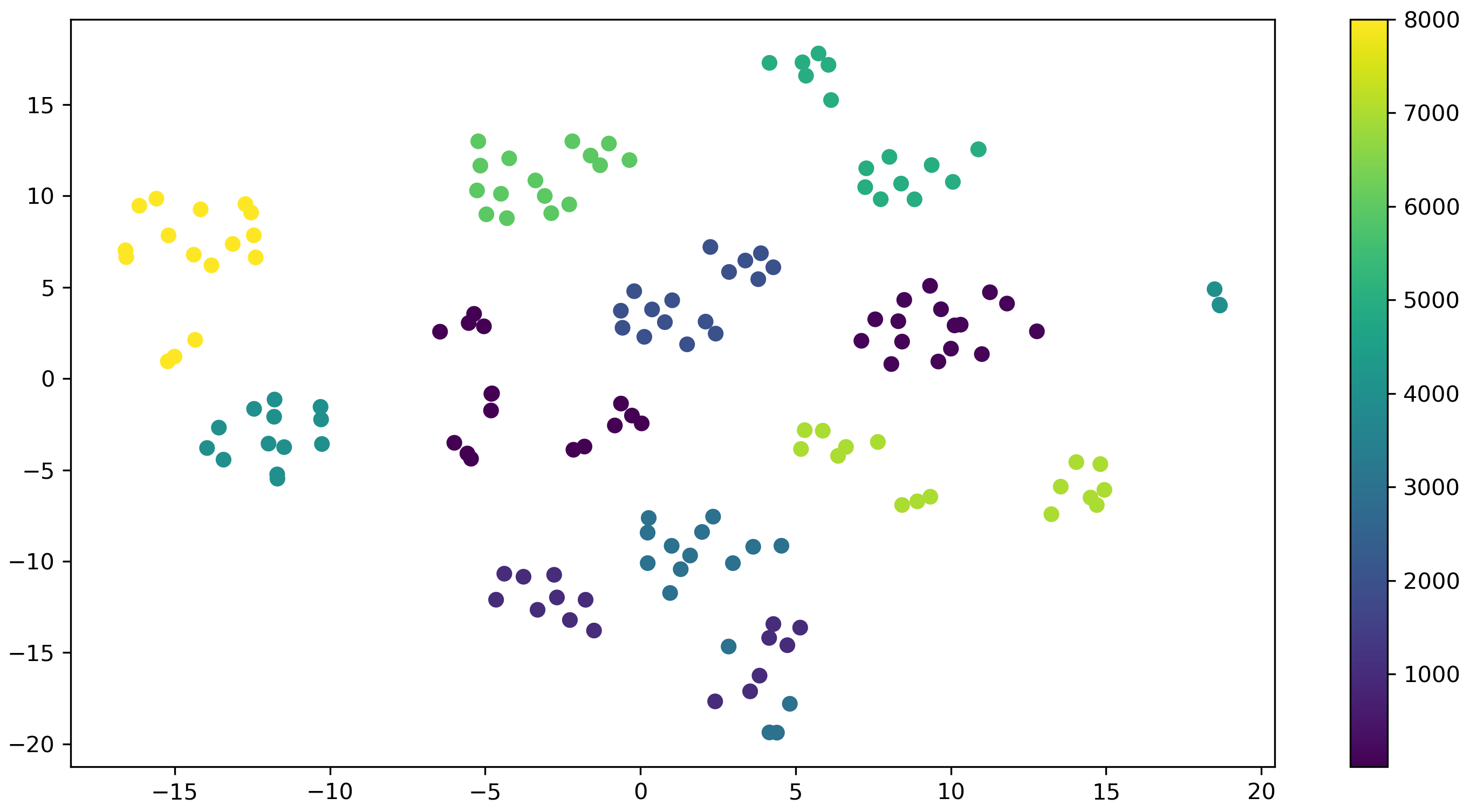}
\caption{SoftMax-10ID}
\label{fig:softmax_dist}
\end{subfigure}
\begin{subfigure}{0.23\textwidth}
\includegraphics[width=\linewidth]{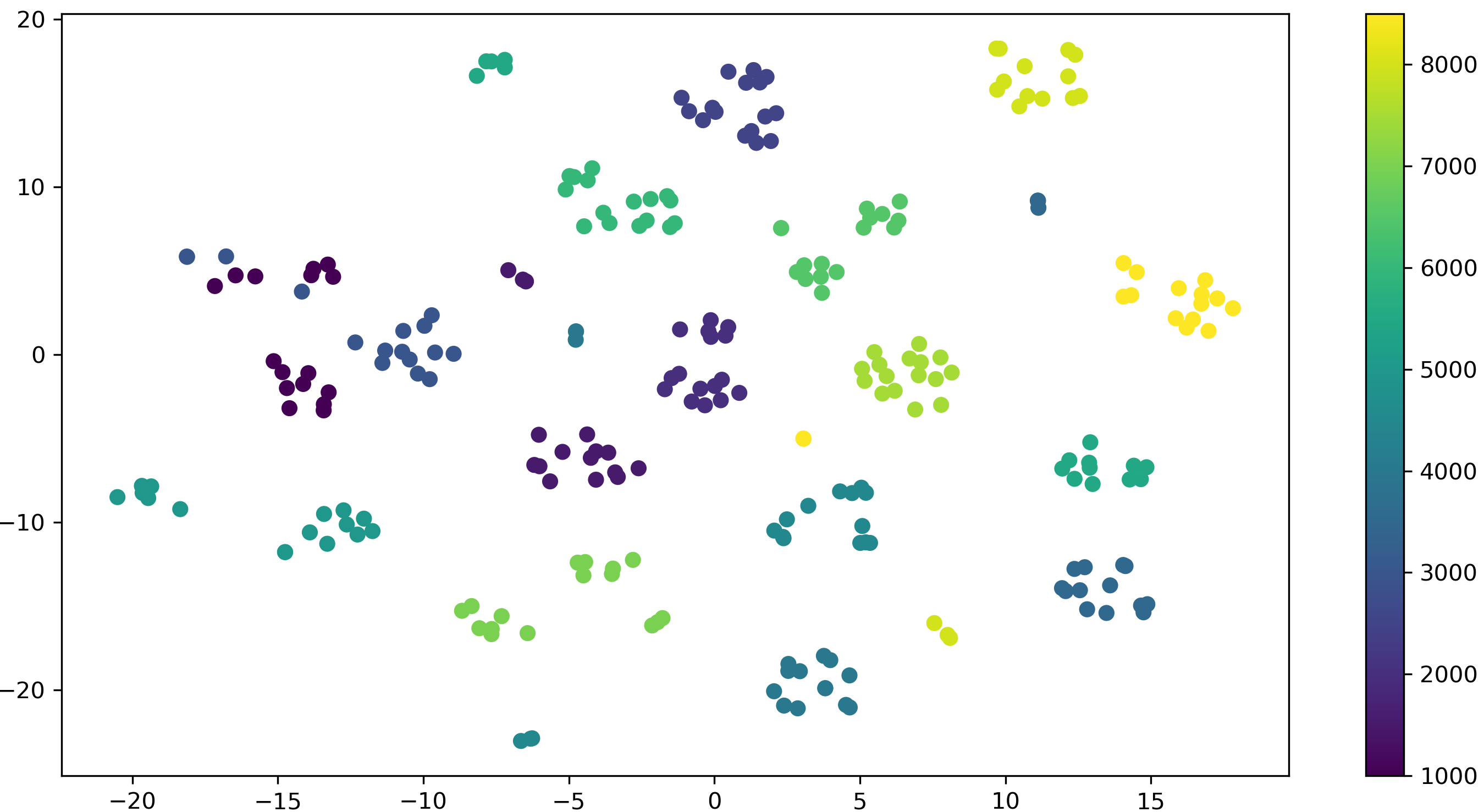}
\caption{SoftMax-16ID}
\label{fig:softmax_16_dist}
\end{subfigure}
\begin{subfigure}{0.23\textwidth}
\includegraphics[width=\linewidth]{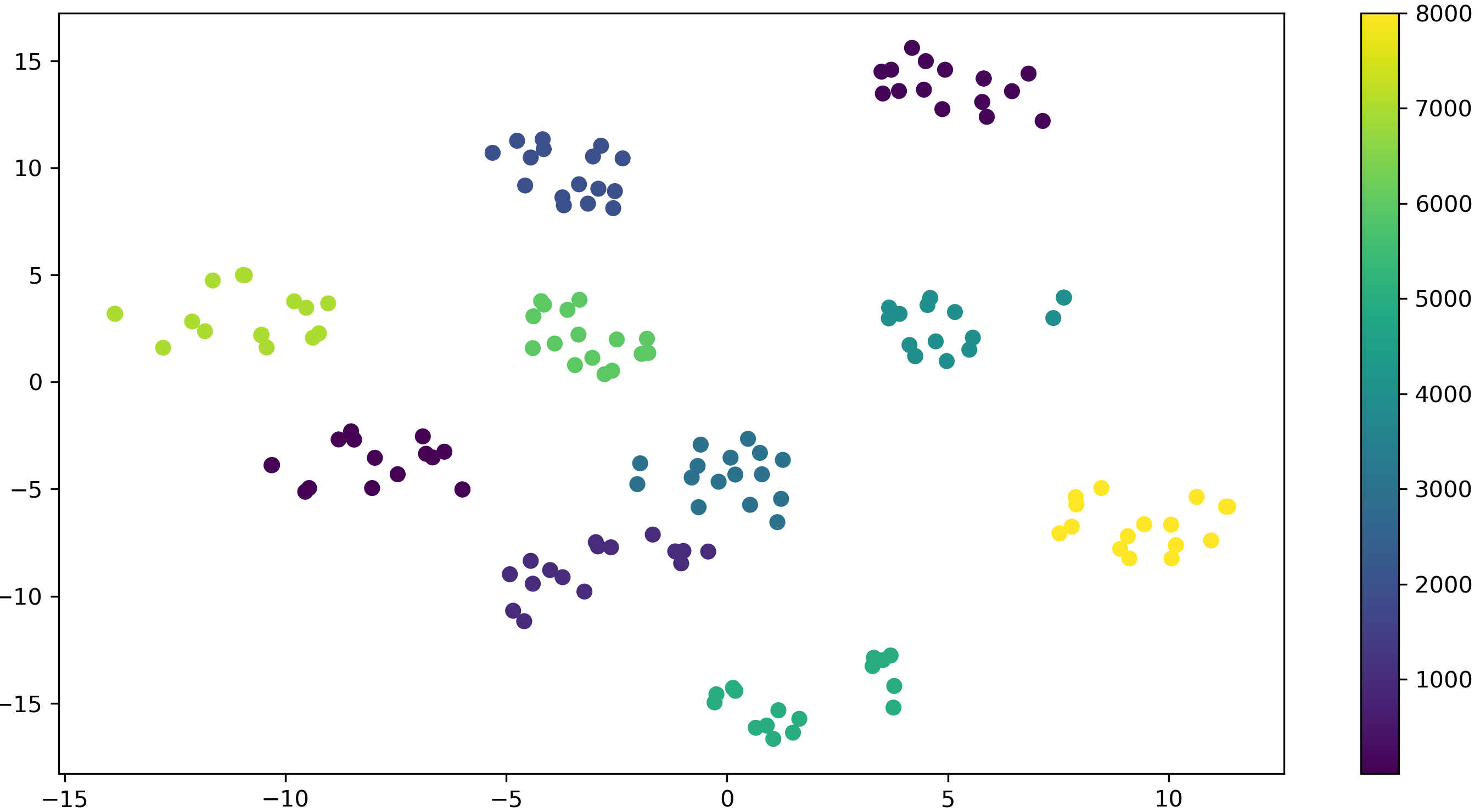}
\caption{ArcFace-10ID}
\label{fig:arc_dist}
\end{subfigure}
\begin{subfigure}{0.23\textwidth}
\includegraphics[width=\linewidth]{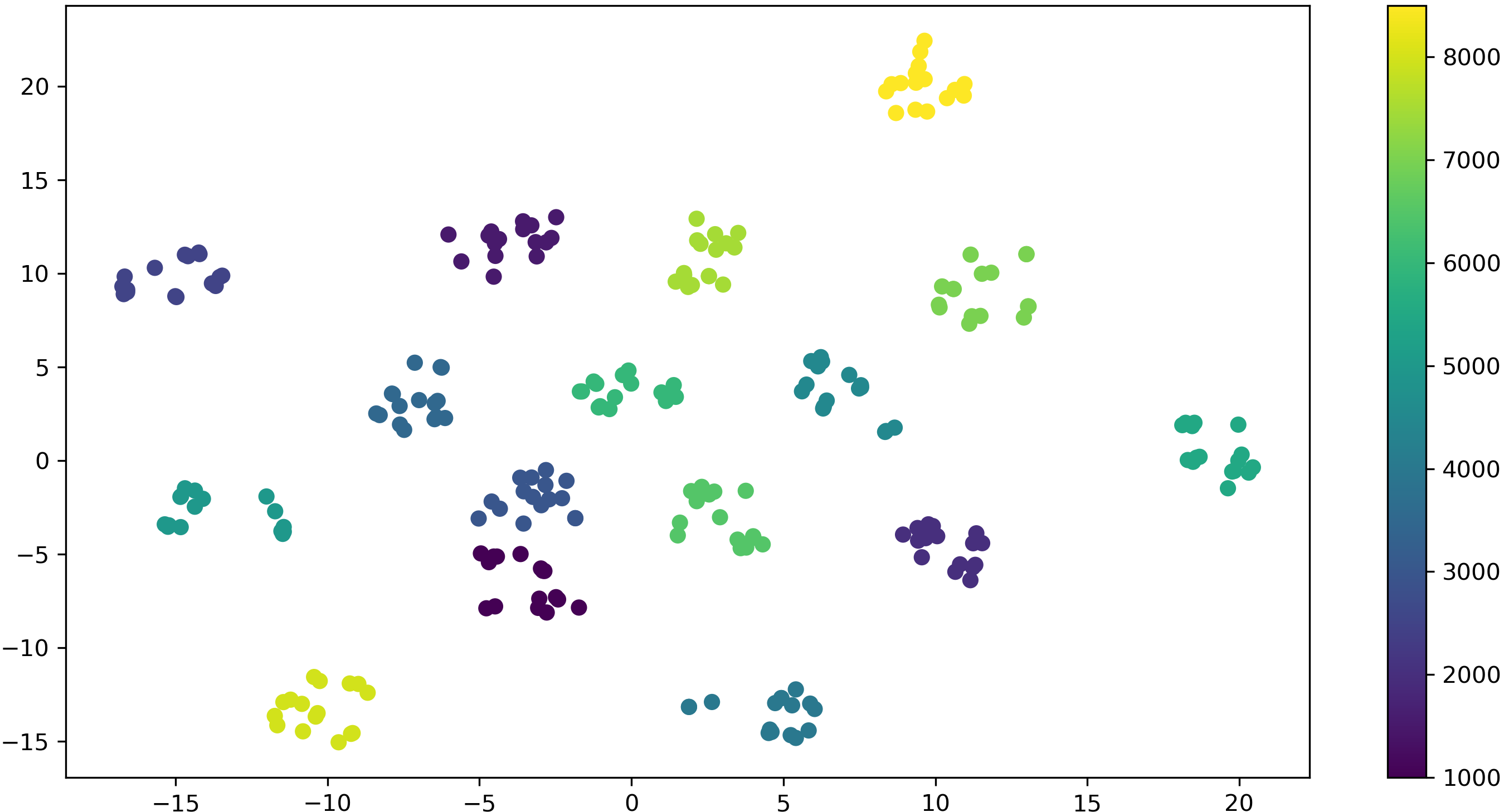}
\caption{ArcFace-16ID}
\label{fig:arc_16_dist}
\end{subfigure}
\begin{subfigure}{0.23\textwidth}
\includegraphics[width=\linewidth]{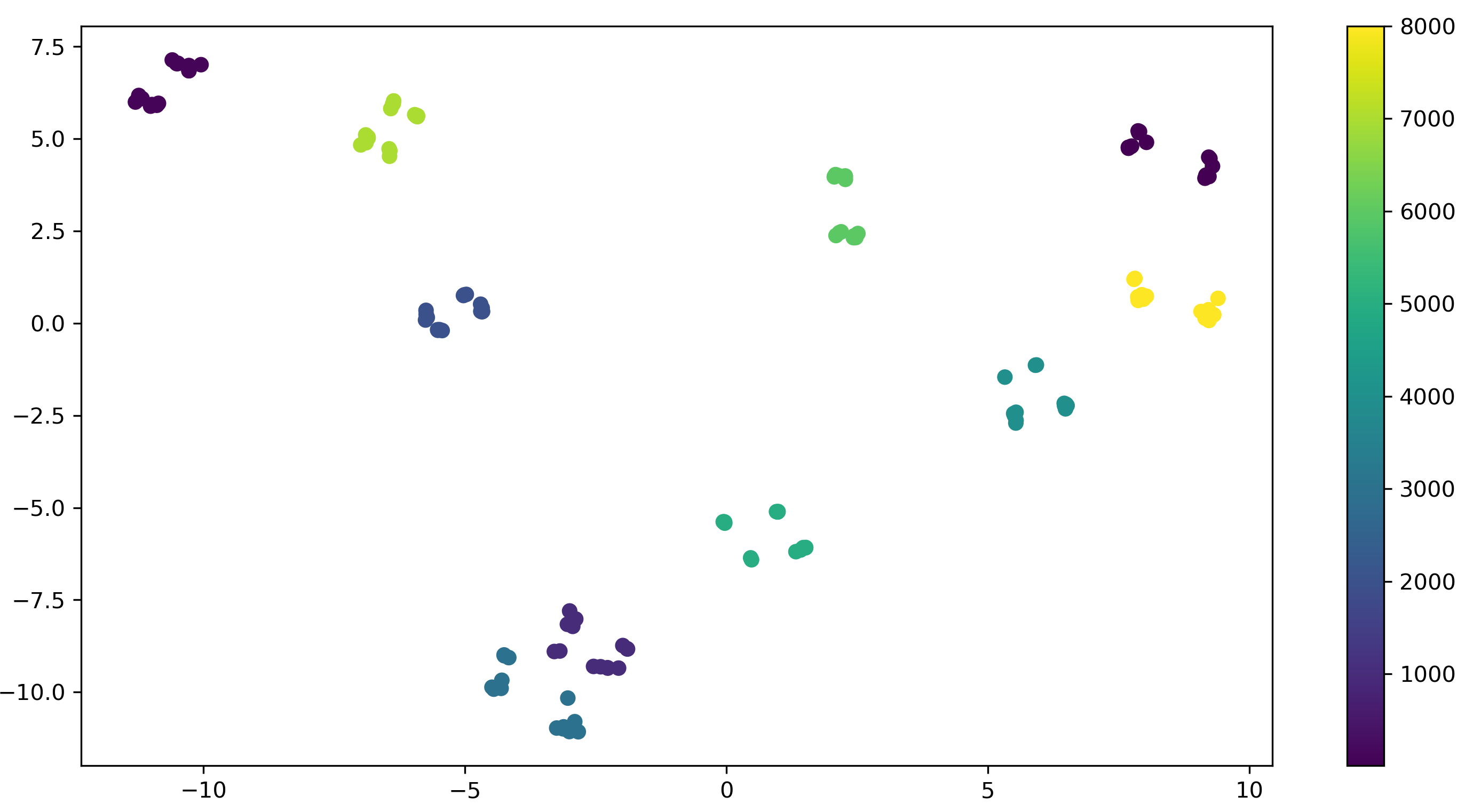}
\caption{DSAM-10ID}
\label{fig:soft_d_dist}
\end{subfigure}
\begin{subfigure}{0.23\textwidth}
\includegraphics[width=\linewidth]{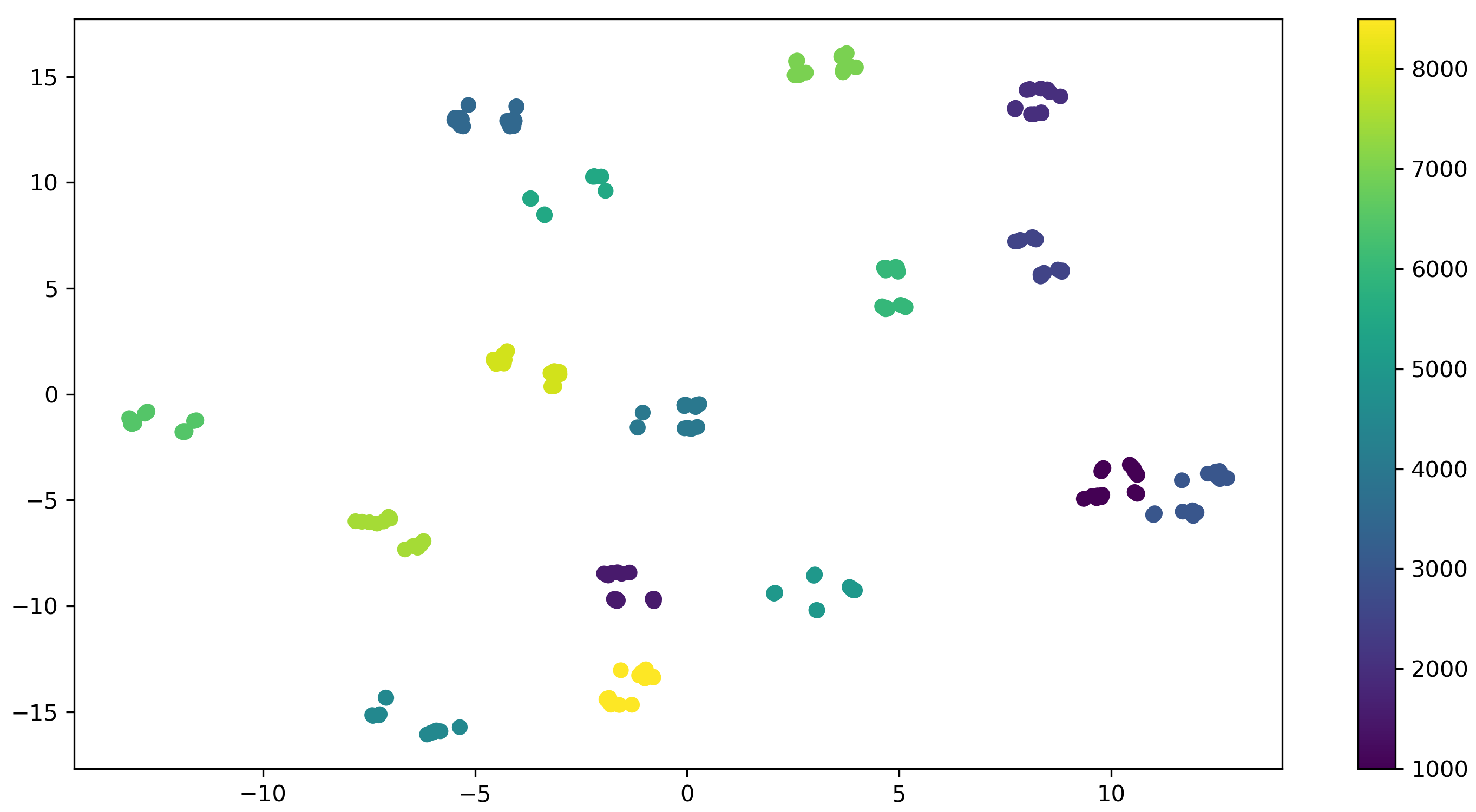}
\caption{DSAM-16ID}
\label{fig:soft_d_16_dist}
\end{subfigure}
%\captionsetup{font={footnotesize}}
\caption{We train three models by different loss functions and we random pick different numbers of IDs from the training set to visualize the distributions of samples belong to different IDs. In the first row, we random choose $10$ IDs and $20$ samples per ID, our DSAM obtains the largest margin between different classes. The second row, we choose the $16$ ids and have $20$ samples in per id, the DSAM also gets the largest interclass distance.}
\label{fig:tSNE}
\end{figure}

Then, we visualize the features learned by different loss functions in the OFS using T-SNE~\cite{maaten2008visualizing}. Fig.~\ref{fig:tSNE} shows that our proposed loss function effectively enhances the intraclass compactness, and more precise boundaries are produced.

\section{Experiments}
\subsection{Experimental Settings}
\textbf{Dataset.}
As shown in Table~\ref{tabel:databases}, we use four challenging and commonly used datasets, PKU-VD (including VD1 and VD2)~\cite{yan2017exploiting}, VehicleID (PKU-VehicleID)~\cite{CVPR16DRDLVehicleReID} and VeRi-776~\cite{liu2016large,ECCV16ProgessiveVehicleReID,liu2017provid}.
\begin{table}[h]
\tabcolsep=3pt
%\footnotesize
\centering
\begin{tabular}{c|c|c|c}
\toprule
Database &Query ID &Gallery ID &Gallery Images \\ \hline
\multirow{3}{*}{PKU-VD1} & \multirow{3}{*}{2000} &18000 (Small) &104887 \\ \cline{3-4}
& &71165 (Medium) &422032 \\ \cline{3-4}
& &71166 (Large) &673323 \\ \hline
\multirow{3}{*}{PKU-VD2} & \multirow{3}{*}{2000} &12000 (Small) &103550 \\ \cline{3-4}
& &40144 (Medium)	&345910 \\ \cline{3-4}
& &40145 (Large) &462652 \\ \hline
\multirow{6}{*}{VehicleID} &13164 &13164 (Q-13164) &13164 \\ \cline{2-4}
&6000 &6000 (Q-6000) &6000 \\ \cline{2-4}
&3200 &3200 (Q-3200) &3200 \\ \cline{2-4}
&2400 &2400 (Q-2400) &2400 \\ \cline{2-4}
&1600	&1600 (Q-1600) &1600 \\ \cline{2-4}
&800 &800 (Q-800) &800 \\ \hline
VeRi-776 &200 &200 (VeRi-test) &11579 \\
\bottomrule
\end{tabular}
%\captionsetup{font={footnotesize}}
\caption{The vehicle ReID datasets used for training and testing. The names in the parentheses are used to denote the testing sets in our experiments.}
\label{tabel:databases}
\end{table}
Each dataset in PKU-VD is provided with three testing sets and one training set, and the testing sets of different sizes have the same number of IDs for the query, but the IDs in the gallery are different. For the VehicleID dataset, the number of query IDs is the same as that of gallery IDs. VeRi-776 is a small database with only 776 vehicles, but this dataset has many vehicle information, which can be used as auxiliary information for model training.

\textbf{Implementation Details.}
In all the experiments, the input images resized to $256 \times 256$ pixels before sending them to the network. We use the ImageNet pre-trained ResNet-50~\cite{he2016deep} as the basic network architecture by replacing the last fully connected layer with a BN-Dropout structure. To avoid losing too much information, we set the stride of the last convolutional layer to 1. For the ArcFace~\cite{deng2019arcface} baseline, which is the state-of-the-art loss of all angular margin losses, we set the embedding size to $2048$, m to $0.5$, and s to $64$ following the settings in~\cite{deng2019arcface}.
\begin{table}
\tabcolsep=3pt
%\footnotesize
\centering
\begin{tabular}{c|ccc}
%\hline
\toprule
Method &mAP	&cmc1 &cmc5  \\ \hline
R50+SoftMax+DSAM(0.7) &65.37\% &90.16\% &96.23\% \\ \hline
R50+SoftMax+DSAM(0.8) &65.89\% &90.71\% &96.62\% \\ \hline
R50+SoftMax+DSAM(0.9) &\textbf{66.20\%} &\textbf{90.92\%} &\textbf{96.87\%} \\
%\hline
\bottomrule
\end{tabular}
%\captionsetup{font={footnotesize}}
\caption{We use DSAM with different margins and combine them with the SoftMax loss under the backbone network ResNet-50. The results are evaluated on the VeRi-776 dataset.}
\label{tabel:margin}
\end{table}

\begin{table}[h]
\small
\centering
\tabcolsep=4pt
\begin{tabular}{c|c|c|c|c|c}
%\hline
\toprule
\multirow{2}{*}{Method} &\multicolumn{5}{|c}{VehicleID, mAP\%} \\ \cline{2-6}
&Q-13164 &Q-6000 &Q-3200 &Q-2400 &Q-1600 \\ \hline
S(baseline)	&59.88\% &63.61\% &64.64\% &67.36\% &68.68\% \\ \hline
S+D	&\textbf{66.95\%} &\textbf{71.38\%} &\textbf{74.10\%} &\textbf{76.70\%}	&\textbf{78.88\%} \\
S+T	&62.71\% &67.30\% &69.15\% &72.24\% &73.88\% \\
NS	&49.49\% &54.55\% &56.70\% &59.19\%	&61.61\% \\
NS+D &59.73\% &65.44\% &68.25\%	&71.24\% &73.96\% \\
NS+T &56.38\% &61.36\% &63.08\% &65.66\% &68.69\% \\
Arc	&63.24\% &66.97\% &67.39\% &70.09\%	&70.29\% \\
Arc+D &64.25\% &67.36\% &68.67\% &71.32\% &71.88\% \\
Arc+T &64.20\% &67.50\% &68.10\% &70.85\% &71.25\% \\
\bottomrule
%\hline
\end{tabular}
%\captionsetup{font={footnotesize}}
\caption{The ``S" means the baseline SoftMax, ``D" means our DSAM, the ``NS" means the Normalized SoftMax, and ``Arc" means the ArcFace. The results are obtained by training ResNet-50 with different loss functions on the VehicleID dataset.}
\label{tabel:loss}
\end{table}

\begin{table}[h]
\small
%\captionsetup{font={footnotesize}}
\centering
\tabcolsep=3pt
\begin{tabular}{c|c|c|c|c|c}
%\hline
\toprule
\multicolumn{6}{c}{mAP (\%) in three datasets} \\ \hline
\multirow{2}{*}{Method} &\multicolumn{2}{|c|}{PKU-VD1} &\multicolumn{2}{|c|}{VehicleID} &VeRi-776 \\ \cline{2-6}
&Large &Small &Q-13164 &Q-2400 &VeRi-test \\ \hline
S(baseline)	&65.02\% &85.14\% &59.88\% &67.36\%	&60.07\% \\ \hline
S+D &\textbf{75.43\%} &\textbf{94.60\%} &\textbf{66.95\%} &\textbf{76.70\%} &66.20\% \\
S+T &73.02\% &93.20\% &62.71\% &72.24\% &63.63\% \\
Arc &73.44\% &92.33\% &63.24\% &70.09\% &65.77\% \\
Arc+D &74.81\% &93.95\%	&64.25\% &71.32\% &\textbf{68.48\%} \\
\bottomrule
%\hline
\end{tabular}
\caption{The results of different loss function on three different scale datasets.}
\label{tabel:diff_dataset_loss}
\end{table}
For DSAM, we set margin to $0.9$, $\gamma$ to $0.8$, and $\lambda$ to $0.05$ by cross-validation. We use the batch-hard strategy as the triplet loss in our experiments according to ~\cite{kuma2019vehicle}. Each batch contains $32$ different IDs, and each ID contains $8$ different images. During training, we use the SGD optimizer by setting the momentum to $0.9$ and the weight decay to $0.0005$. The learning rate starts from $0.01$ and is divided by $10$ every $10$ epochs. When the model converges, the learning rate is $0.00001$. During testing, we use the output before the loss function as the feature representations and use the cosine distance to calculate the similarity between the query and gallery images.

\begin{table*}[!t]
\small
\tabcolsep=3pt
\centering
\begin{tabular}{c|c|c|c|c|c|c|c|c}
%\hline
\toprule
\multirow{2}{*}{Method} &\multirow{2}{*}{Publication} &\multirow{2}{*}{Backbone} &\multicolumn{2}{|c|}{VeID Q-800} &\multicolumn{2}{|c|}{VeID Q-1600} &\multicolumn{2}{c}{VeID Q-2400} \\ \cline{4-9}
& & &cmc1 &mAP &cmc1 &mAP &cmc1 &mAP \\ \hline
C2F-Rank~\cite{AAAI18VehicleReID} &AAAI 2018 &GoogLeNet &61.10\% &63.50\% &56.20\% &60.00\% &51.40\% &53.00\% \\ \cline{1-1}
VAMI~\cite{CVPR18ViewpointAwareVehicleReID}	&CVPR 2018 &- &63.12\% &- &52.87\% &- &47.34\% &- \\ \cline{1-1}
AAVER~\cite{khorramshahi2019dual} &ICCV 2019 &ResNet-101 &74.69\% &- &68.62\% &- &63.54\% &- \\ \cline{1-1}
HVE~\cite{lou2019embedding} &TIP 2019 &Resnet50 &75.11\% &77.50\% &71.78\% &74.20\% &69.30\% &71.00\% \\ \cline{1-1}
RAM~\cite{liu2018ram}	&ICME 2018 &VGG	&75.20\% &- &72.30\% &- &67.70\% &- \\ \cline{1-1}
DF~\cite{zheng2019attributes}	&- &ResNet-50 &75.23\% &78.03\% &72.15\% &74.87\% &70.46\% &73.15\% \\ \cline{1-1}
P-R~\cite{he2019part} &CVPR 2019 &ResNet-50 &78.40\% &- &75.00\% &- &74.20\% &- \\ \cline{1-1}
PRN~\cite{Chen_2019_CVPR_Workshops}	&CVPR\_W 2019 &ResNet-50 &78.92\% &- &74.94\% &- &71.58\% &- \\ \cline{1-1}
RNN-HA~\cite{wei2018coarse}	&ACCV 2018 &VGG	&83.80\% &- &81.90\% &- &81.10\% &- \\ \cline{1-1}
PVEN~\cite{meng2020parsing} &CVPR2020 &SeResNeXt50 &84.70\% &- &80.60\% &- &77.80\% &- \\ \cline{1-1}
%BS~\cite{kuma2019vehicle} &- &- &78.80\% &\textbf{86.19\%} &73.41\% &\textbf{81.69\%} &69.33\% &\textbf{78.16\%} \\  \cline{1-1}
VANET~\cite{chu2019vehicle}	&ICCV 2019 &GoogLeNet &88.12\% &- &83.17\% &- &80.35\% &- \\ \cline{1-1} \hline
\multicolumn{2}{c|}{R50+S+D(Ours)} &ResNet-50&\textbf{93.75\%} &\textbf{83.59\%} &\textbf{91.06\%} &\textbf{78.88\%} &\textbf{90.29\%} &\textbf{76.70\%} \\ \cline{1-2}
\multicolumn{2}{c|}{R50+Arc+D(Ours)} &ResNet-50 &92.12\% &74.35\% &89.79\% &71.88\% &89.79\% &71.32\% \\
\bottomrule
%\hline
\end{tabular}
%\captionsetup{font={footnotesize}}
\caption{We compare our results with the recent state-of-the-arts on VehicleID. The CVPR\_W means CVPR Workshop.}
\label{tabel:sota_veid}
\end{table*}

\begin{figure}[t]
\centering
\begin{subfigure}{0.23\textwidth}
\includegraphics[width=\linewidth]{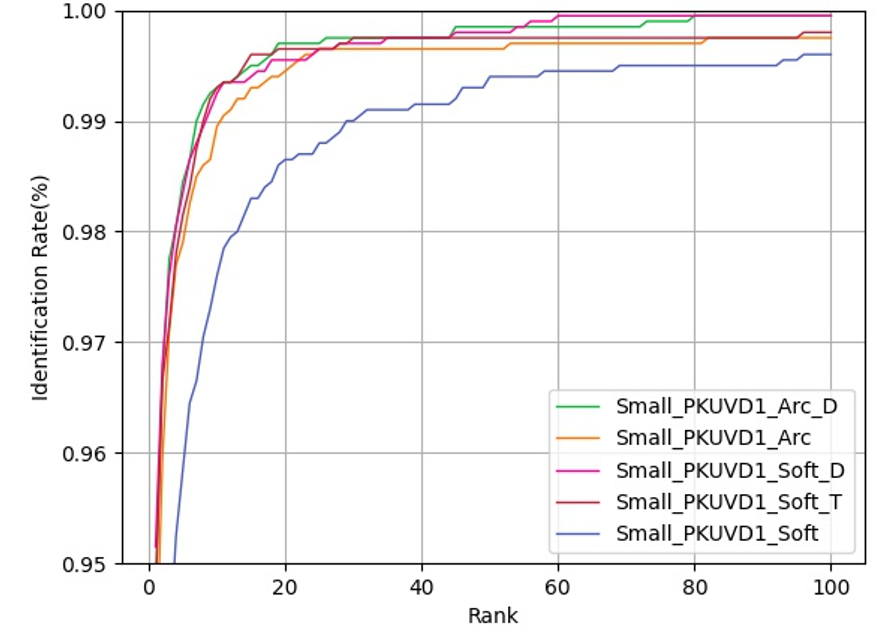}
\caption{PKU-VD1 CMC Curve}
\label{fig:pku_cmc}
\end{subfigure}
\begin{subfigure}{0.23\textwidth}
\includegraphics[width=\linewidth]{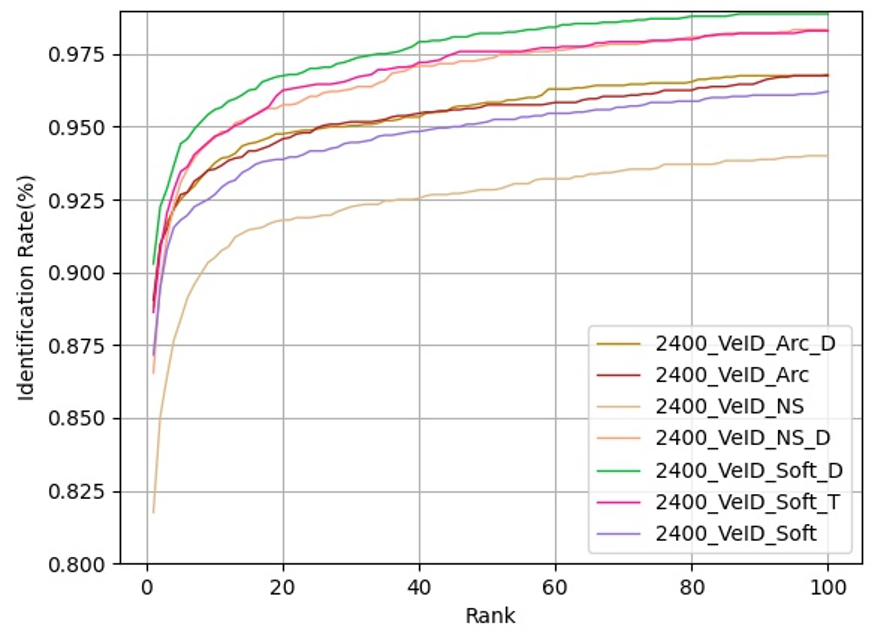}
\caption{PKU-VehicleID CMC Curve}
\label{fig:vehicle_cmc}
\end{subfigure}
%\captionsetup{font={footnotesize}}
\caption{CMC result of different loss function on PKU-VD1’s small testing set and PKU-VehicleID’s Q-2400 testing set}
\label{fig:cmc_curve}
\end{figure}

\textbf{Evaluation Metrics.}
We exploit the widely used mean-average-precision (mAP) and Cumulative Matching Characteristic (CMC-n) to evaluate the performance of different methods on different databases. As for the PKU-VD and VehicleID databases, we also use the CMC curve to analyze different loss functions.

\subsection{Ablation Study}
In Table~\ref{tabel:margin}, we first explore the optimal value of the margin. It can be seen that when $margin = 0.9$, the experimental results are the best, so we set the margin value of DSAM to $0.9$ in all subsequent experiments.

Table~\ref{tabel:loss} shows the experimental results on the VehicleID~\cite{CVPR16DRDLVehicleReID} dataset. Here, we combined DSAM (D) with SoftMax, normalized SoftMax~\cite{liu2017sphereface}, and AraFace~\cite{deng2019arcface}. DSAM boosts the performance of all the SoftMax-based losses, especially the baseline SoftMax (S). The original performance of the Normalized SoftMax (NS) is relatively bad, but by incorporating our DSAM makes the NS+D even surpass the ArcFace (Arc) on some small testing sets. Besides, it shows from multiple testing sets that DSAM is better than the triplet Loss (T)~\cite{CVPR15FaceNet} and make the S+D gain the best score.

We also test our DSAM loss on different datasets, and the results are showed in Table~\ref{tabel:diff_dataset_loss}. We can see that our DSAM (D) works well on both small and large scale testing sets. In contrast, the triplet loss (T) works better on small scale testing sets than on large scale testing sets. On the large testing set of PKU-VD1~\cite{yan2017exploiting}, the S+T’s mAP is lower than Arc’s, but on the small one, the S+T’s mAP is bigger than Arc’s. As for the dataset which has a small scale training set like VeRi-776~\cite{liu2016large,ECCV16ProgessiveVehicleReID,liu2017provid}, the S+T and S+D are not work well as Arc, but our DSAM can still improve the SoftMax-based losses and works better than T. According to all the results from these three datasets, if the training set is small, the Arc+D works the best, and if the training set is large, the S+D works the best.

%% This will show in supplementary material
\begin{table}[t]
\footnotesize
\centering
\tabcolsep=1.5pt
\begin{tabular}{c|c|c|c|c}
%\hline
\toprule
\multirow{2}{*}{Method} &\multirow{2}{*}{PKU} &\multicolumn{1}{|c|}{Large} &\multicolumn{1}{|c|}{Medium} &\multicolumn{1}{c}{Small} \\ \cline{3-5}
& &mAP &mAP &mAP \\ \hline
BW (18*16)~\cite{kuma2019vehicle} &\multirow{4}{*}{VD1} &58.77\% &67.28\% &87.48\% \\ \cline{1-1}
ATTS+MGLR~\cite{yan2017exploiting} & &51.10\% &58.30\% &79.10\% \\ \cline{1-1} \cline{3-5}
R50+S+D (Ours) & &\textbf{75.43\%} &\textbf{85.00\%} &\textbf{94.60\%} \\ \cline{1-1}
R50+Arc+D (Ours) & &74.81\% &84.15\% &93.95\% \\ \hline
BW (18*16)~\cite{kuma2019vehicle} &\multirow{4}{*}{VD2} &63.63\% &69.87\% &84.55\% \\ \cline{1-1}
ATTS+MGLR~\cite{yan2017exploiting} & &55.30\% &60.60\% &74.70\% \\ \cline{1-1} \cline{3-5}
R50+Arc+D(Ours) & &\textbf{86.71\%} &\textbf{88.64\%} &\textbf{92.30\%} \\ \cline{1-1}
R50+S+D(Ours) & &85.96\% &88.43\% &92.75\% \\
\bottomrule
%\hline
\end{tabular}
%\captionsetup{font={footnotesize}}
\caption{We compare our results with the state-of-the-arts on PKU-VD.}
\label{tabel:sota_pkuvd}
\vspace{-0.2cm}
\end{table}

For a more intuitive comparison of these loss functions, we plot the CMC-curve of PKU-VehicleID’s~\cite{ECCV16ProgessiveVehicleReID} Q-2400 testing set and the PKU-VD1’s~\cite{yan2017exploiting} small testing set in Figure~\ref{fig:cmc_curve}. S+D shows the superiority over other loss functions in the large scale datasets.
\begin{table}[h]
\small
\centering
\tabcolsep=4pt
\begin{tabular}{c|c|c|c}
%\hline
\toprule
\multicolumn{4}{c}{VeRi-776} \\ \hline
Method &mAP	&cmc1 &cmc5 \\ \hline
PROVID$^*$~\cite{ECCV16ProgessiveVehicleReID}	&27.77\%	&61.44\%	&78.78\% \\
OIFE$^{\dagger}$~\cite{ICCV17OIFEVehicleReID}	&48.00\%	&89.43\%	&-	\\
VAMI$^{\dagger \ddag}$~\cite{CVPR18ViewpointAwareVehicleReID}	&50.13\%	&77.03\%	&90.82\% \\
Path-LSTM$^*$~\cite{shen2017learning}	&58.27\%	&83.49\%	&90.04\% \\
BS~\cite{kuma2019vehicle}	&67.55\%	&90.23\%	&96.42\% \\ \hline
R50+S+D(Ours)	&66.20\%	&90.92\%	&96.87\% \\
R50+Arc+D(Ours)	&\textbf{68.48\%}	&\textbf{91.37\%}	&\textbf{97.14\%} \\
\bottomrule
%\hline
\end{tabular}
%\captionsetup{font={footnotesize}}
\caption{We compare our results with the state-of-the-arts on VeRi-776. ($^*$) indicates the usage of spatio-temporal information.($^{\dagger}$) indicates the usage of additional annotations besides ID label. ($^{\ddag}$) indicates the usage of multi-view information}
\label{tabel:sota_veri}
\vspace{-0.2cm}
\end{table}

\subsection{Comparison with the State-of-the-Art Methods}
% This will show in supplementary material
We compare our result with the state-of-the-arts in Tabel~\ref{tabel:sota_pkuvd}. The model trained by SoftMax with our DSAM obtains the best results on all the testing sets of PKU-VD1~\cite{yan2017exploiting}. On PKU-VD2~\cite{yan2017exploiting}, the model trained by ArcFace~\cite{deng2019arcface} our DSAM obtains the best results in small testing sets. Since the training set of PKU-VD2 is smaller than that of the PKU-VD1, we conclude that ArcFace+DSAM works better than the SoftMax+DSAM in small dataset.

We also compare our results with the state-of-the-art on PKU-VehicleID~\cite{CVPR16DRDLVehicleReID} and VeRi-776~\cite{liu2016large,ECCV16ProgessiveVehicleReID,liu2017provid}, and the experimental results are show  in Tabel~\ref{tabel:sota_veid} and Table~\ref{tabel:sota_veri} respectively. We refer readers to read our supplementary material for more detailed quantitative evaluations on PKU-VD~\cite{yan2017exploiting} dataset. In Tabel~\ref{tabel:sota_veid}, we can see that we get the best score of CMC1 on three testing sets. What is more, the other methods in Tabel~\ref{tabel:sota_veid} have much more complicated model architectures than ours, and we use the ResNet-50 backbone and the SoftMax combined with our DSAM. 
In the Tabel~\ref{tabel:sota_veid}, we can also confirm that the DSAM works better on the large scale datasets than on the small datasets. Nevertheless, the DSAM gets better results on all datasets than other models or loss functions, especially CMC1. From the results in Table~\ref{tabel:sota_veri}, our R50+Arc+D gets the best score in CMC1 and CMC5 and we only use the ID information in VeRi-776~\cite{liu2016large,ECCV16ProgessiveVehicleReID,liu2017provid}.

\section{Conclusions}
In this paper, we proposed a Distance Shrinking with Angular Marginalizing loss function, termed as DSAM, which performs hybrid learning in both the Original Feature Space (OFS) and the Feature Angular Space (FAS) by using the local verification and the global identification information respectively. It shrinks the distance between samples of the same class locally in the OFS, while keeps samples of different classes far away in the FAS. Also, it does not introduce any additional learning parameters and is very easy to implement, which can be integrated with all the existing SoftMax based loss functions to boost their performances. Extensive experimental analyses and comparisons with many competing methods on three large vehicle ReID benchmarks demonstrate the proposed loss function's effectiveness. In addition, the superior performance of shrinking the distance between samples of the same class locally in the OFS and the ability to keep instances of different classes far away in the FAS features further suggested that DSAM loss may improve the work on the person ReID task.

\section{\textbf{Supplementary Material}}
\section{Databases and Setting}
\subsection{Samples of variegated databases}
\begin{figure}[!ht]
\begin{subfigure}{0.45\textwidth}
\centering
\includegraphics[width=\linewidth]{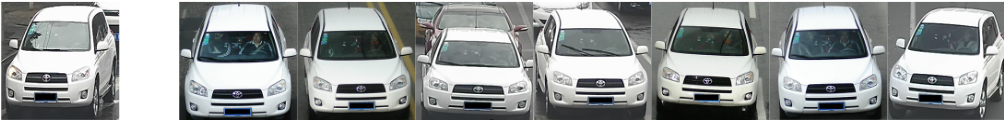}
\captionsetup{justification=centering}
%\small
\caption{PKU-VD1}
\label{fig:pku-vd1}
\end{subfigure}
\begin{subfigure}{0.45\textwidth}
\centering
\includegraphics[width=\linewidth]{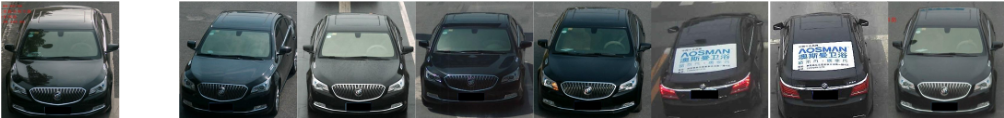}
\captionsetup{justification=centering}
%\small
\caption{PKU-VehicleID}
\label{fig:veid}
\end{subfigure}
\begin{subfigure}{0.45\textwidth}
\centering
\includegraphics[width=\linewidth]{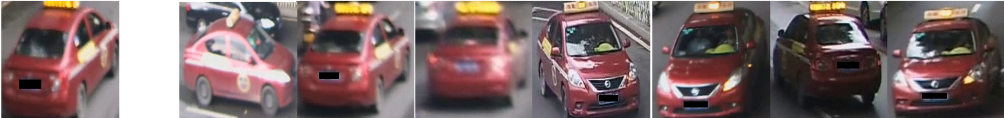}
\captionsetup{justification=centering}
%\small
\caption{VeRi-776}
\label{fig:veri776}
\end{subfigure}
\begin{subfigure}{0.45\textwidth}
\centering
\includegraphics[width=\linewidth]{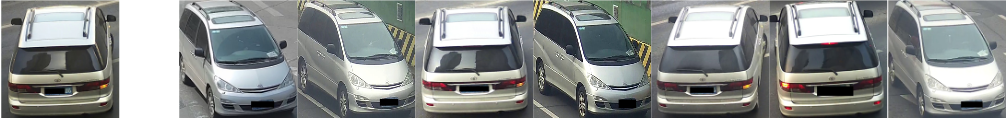}
\captionsetup{justification=centering}
%\small
\caption{Veri-Wild}
\label{fig:wild}
\end{subfigure}
\begin{subfigure}{0.45\textwidth}
\centering
\includegraphics[width=\linewidth]{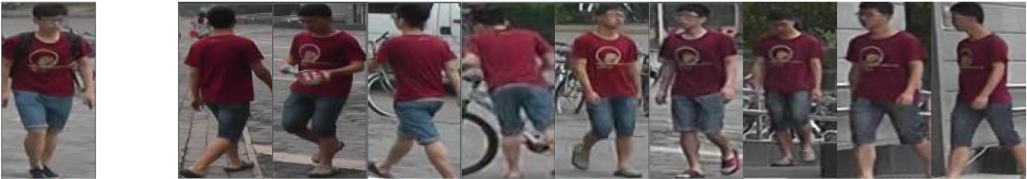}
\captionsetup{justification=centering}
%\small
\caption{Market-1501}
\label{fig:market}
\end{subfigure}
\begin{subfigure}{0.45\textwidth}
\centering
\includegraphics[width=\linewidth]{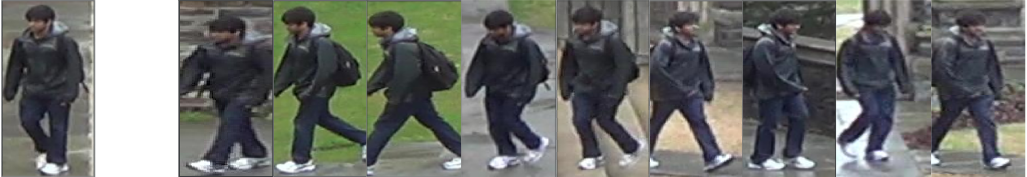}
\captionsetup{justification=centering}
%\small
\caption{DukeMTMC-reID}
\label{fig:duke}
\end{subfigure}
\begin{subfigure}{0.45\textwidth}
\centering
\includegraphics[width=\linewidth]{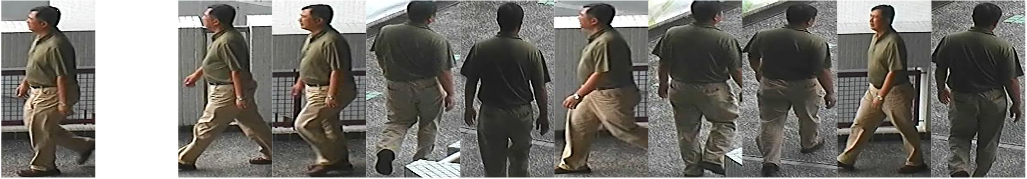}
\captionsetup{justification=centering}
%\small
\caption{CUHK03}
\label{fig:cuhk}
\end{subfigure}
\centering
\caption{The samples of four different vehicle databases and three person databases. The image in left of the blank is the query image and the images in right of the blank are gallery images.}
\label{fig:data_samples}
\vspace{-0.2cm}
\end{figure}

\begin{table}[!ht]
\footnotesize
\tabcolsep=0.5pt
\centering
\begin{tabular}{c|c|c|c|c}
%\hline
\toprule
Type &Database &Query ID &Gallery ID &Gallery Images \\ \hline
\multirow{3}{*}{Vehicle} &Veri-Wild (Small) &3000 &3000 &38861 \\ \cline{2-5}
&Veri-Wild (Medium) &5000 &5000 &64389 \\ \cline{2-5}
&Veri-Wild (Large) &10000 &10000 &128517 \\ \hline \hline
\multirow{4}{*}{Person} &Market-1501 &750 &751 &15913 \\ \cline{2-5}
&DukeMTMC-reID &702 &1110 &17661 \\ \cline{2-5}
&CUHK03 (Labeled) &700 &700 &5328 \\ \cline{2-5}
&CUHK03 (Detected) &700 &700 &5332 \\
%\hline
\bottomrule
\end{tabular}
\caption{The vehicle ReID datasets and person ReID datasets used for training and testing. The names in the parentheses denote the different testing sets in our experiments.}
\label{table:databases}
\vspace{-0.2cm}
\end{table}
To better show the challenge of Vehicle ReID, we randomly pick up some samples from vehicle databases used in this paper. As Fig.~\ref{fig:data_samples} shows, vehicles simultaneously exhibit large intraclass viewpoint variations and interclass visual similarities. Different samples with the same id have large intraclass variations (e.g., different backgrounds, viewpoints, illuminations, etc.), which indicate the margin between different classes are small, and many hard instances per class.
%\vspace{-0.2cm}
\begin{table}[!ht]
\footnotesize
\centering
\tabcolsep1pt
\begin{tabular}{c|c|c|c|c|c|c}
%\hline
\toprule
\multirow{3}{*}{Method} &\multicolumn{6}{|c}{Veri-Wild} \\ \cline{2-7}
&\multicolumn{2}{|c}{Small} &\multicolumn{2}{|c}{Medium} &\multicolumn{2}{|c}{Large} \\ \cline{2-7}
&mAP &CMC-1 &mAP &CMC-1 &mAP &CMC-1 \\ \hline
S(baseline)	&51.40\% &89.47\% &47.04\% &86.19\% &41.42\% &81.78\% \\ \hline
S+D	&\textbf{69.24\%} &\textbf{89.90\%} &\textbf{63.50\%} &\textbf{86.20\%} &\textbf{50.55\%} &\textbf{81.80\%} \\ 
S+T	&53.28\% &89.70\% &48.31\% &85.90\% &42.00\% &81.38\% \\ 
Arc	&49.11\% &88.77\% &- &- &- &- \\ 
Arc+D &54.67\% &89.70\%	&49.23\% &85.08\% &42.65\% &80.80\% \\ 
\bottomrule
%\hline
\end{tabular}
\caption{The ``S" means the baseline SoftMax, ``D" means our DSAM loss, ``Arc" means the ArcFace loss. The results are obtained by training ResNet-50 with different loss functions on the Veri-Wild dataset.}
\label{table:loss_wild}
\vspace{-0.2cm}
\end{table}
\begin{table*}[!ht]
\centering
%\footnotesize
%\tabcolsep0.5pt
\begin{tabular}{c|c|c|c|c|c|c}
%\hline
\toprule
\multicolumn{7}{c}{Veri-Wild} \\ \hline
\multirow{2}{*}{Method} &\multicolumn{2}{c|}{Small} &\multicolumn{2}{c|}{Medium} &\multicolumn{2}{c}{Large} \\ \cline{2-7}
&mAP &cmc1 &mAP &cmc1 &mAP &cmc1 \\ \hline
GoogLeNet~\cite{yang2015large} &24.27\% &57.16\% &24.15\% &53.16\% &21.53\% &44.61\% \\ \cline{1-1}
CCL~\cite{CVPR16DRDLVehicleReID} &22.50\% &56.96\% &19.28\% &51.92\% &14.81\% &44.60\% \\ \cline{1-1}
HDC~\cite{yuan2017hard} &29.14\% &57.10\% &24.76\% &49.64\% &18.30\% &43.97\% \\ \cline{1-1}
U-GAN~\cite{zhu2017unpaired} &29.86\% &58.06\% &24.71\% &51.58\%	&18.23\% &43.63\% \\ \cline{1-1}
FDA-Net~\cite{lou2019veri}	&35.11\% &64.03\% &29.80\% &57.82\% &22.78\% &49.43\% \\ \cline{1-1}
MLSL~\cite{alfasly2019multi}	&46.32\% &86.03\% &42.37\% &83.00\% &36.61\% &77.51\% \\ \cline{1-1}
BS~\cite{kuma2019vehicle}	&68.79\% &82.90\% &61.11\% &77.68\% &49.79\% &69.59\% \\ \hline
R50+S+D(Our) &\textbf{69.24\%} &\textbf{89.90\%} &\textbf{63.50\%} &\textbf{86.20\%} &\textbf{50.55\%}	&\textbf{81.80\%} \\
\bottomrule
%\hline
\end{tabular}
\caption{We compare our results with the recent state-of-the-arts on Veri-Wild.}
\label{table:sota_wild}
\vspace{-0.2cm}
\end{table*}
\begin{table*}[!ht]
%\footnotesize
%\tabcolsep=2pt
\centering
\begin{tabular}{c|c|c|c|c|c}
%\hline
\toprule
\multirow{2}{*}{Method} &\multirow{2}{*}{Publication}  &\multicolumn{2}{c}{DukeMTMC-reID} &\multicolumn{2}{|c}{Maarket-1501} \\ \cline{3-6}
& &mAP(\%) &R-1(\%) &mAP(\%) &R-1(\%) \\ \hline
PGFA~\cite{miao2019pose} &ICCV'19 &65.5 &82.6 &76.8 &91.2 \\
CAMA~\cite{yang2019towards} &CVPR'19 &72.9 &85.8 &84.5 &94.7 \\
$P^2-Net$~\cite{guo2019beyond} &ICCV'19 &73.1 &86.5 &85.6 &95.2 \\
SNR~\cite{jin2020style} & CVPR'20 &73.2 &85.5 &82.3 &93.4 \\
IANet~\cite{hou2019interaction} &CVPR'19 &73.4 &87.1 &83.1 &94.4 \\
OSNet~\cite{zhou2019omni} \textbf{(B)} &ICCV'19 &73.5 &86.6 &84.9 &94.8 \\
DSA-re-ID~\cite{zhang2019densely} &CVPR'19 &74.3 &86.2 &87.6 &95.7 \\
AANet~\cite{tay2019aanet}& CVPR'19 &74.3 &87.7 &83.4 &93.9 \\
DGNet~\cite{zheng2019joint}&CVPR'19 &74.8 &86.6 &86.0 &94.8 \\
CtF~\cite{wang2020faster} &ECCV'20 &74.8 &87.6 &84.0 &93.7 \\
HOReID~\cite{wang2020high} &CVPR'20 &75.6 &86.9 &84.9 &94.2 \\
SCSN~\cite{chen2020salience} &CVPR'20 &79.0 &91.0 &88.5 &95.7 \\
ISP~\cite{zhu2020identity} &ECCV'20 &80.0 &89.6 &88.6 &95.3 \\ \hline
OSNet+S+T (w/o rr) &-  &76.6 &88.1 &86.1 &94.6 \\
OSNet+S+D (w/o rr) &Ours &78.6 &89.2 &87.7 &95.3 \\
OSNet+S+D (rr) &Ours &\textbf{88.9} &\textbf{91.3} &\textbf{94.4} &\textbf{96.1}\\
\bottomrule
%\hline
\end{tabular}
%\captionsetup{font={footnotesize}}
\caption{Comparison with the recent state-of-the-art on DukeMTMC-reID datasets. ``\textbf{(B)}" means baseline. ``rr" means re-ranking~\cite{zhong2017re}, ``D" denotes the DSAM loss and ``S" means SoftMax.}
\label{table:sota_duke}
%\vspace{-0.5cm}
\vspace{-0.2cm}
\end{table*}
\begin{table*}[!ht]
%\footnotesize
%\tabcolsep=2pt
\centering
\begin{tabular}{c|c|c|c|c|c}
%\hline
\toprule
\multicolumn{6}{c}{CUHK03} \\ \hline
\multirow{2}{*}{Method} &\multirow{2}{*}{Publication} &\multicolumn{2}{c|}{Labeled} &\multicolumn{2}{c}{Detected} \\ \cline{3-6}
& &mAP(\%) &R-1(\%) &mAP(\%) &R-1(\%) \\ \hline
MLFN~\cite{chang2018multi} &CVPR'18 &49.2 &54.7 &47.8 &52.8 \\
MGN~\cite{wang2018learning} &MM'18 &67.4 &68.0 &66.0 &68.0 \\
MHN~\cite{chen2019mixed} &ICCV'19 &72.4 &77.2 &65.4 &71.7 \\
Auto-ReID~\cite{quan2019auto} &ICCV'19 &73.0 &77.9 &69.3 &73.3 \\
BDB~\cite{dai2019batch} &ICCV'19 &76.7 &79.4 &73.5 &76.4 \\
Pyramid~\cite{zheng2019pyramidal} &CVPR'19 &76.9 &78.9 &74.8 &78.9 \\
OSNet~\cite{zhou2019omni} \textbf{(B)} &ICCV'19 &- &- &67.8 &72.3 \\ \hline
OSNet+S+T (w/o rr) &-  &72.4 &75.7 &69.1 &73.1 \\
OSNet+S+D (w/o rr) &Ours &76.7 &78.4 &74.4 &77.3 \\
OSNet+S+D (rr) &Ours &\textbf{87.9} &\textbf{86.4} &\textbf{85.2} &\textbf{83.6} \\
\bottomrule
%\hline
\end{tabular}
%\captionsetup{font={footnotesize}}
\caption{Comparison with the recent state-of-the-art on CUHK03 (Labeled) and CUHK03 (Detected) datasets. ``\textbf{(B)}" means baseline. ``rr" means re-ranking~\cite{zhong2017re}, ``D" denotes the DSAM loss and ``S" means SoftMax.}
\label{table:sota_cuhk}
\vspace{-0.2cm}
%\vspace{-0.5cm}
\end{table*}
So the angular margin losses based on the SoftMax loss may not be useful. In order to show DSAM loss outperforms the angular margin loss and triplet loss in the large-intraclass-distance Vehicle ReID tasks, we experiment on Veri-Wild~\cite{lou2019veri} database, which has considerable intraclass variations. As Fig.~\ref{fig:wild} shows, samples in Veri-Wild database with the same id are under many different viewpoints, different illuminations, and different backgrounds. Though samples in VeRi-776~\cite{liu2016large,ECCV16ProgessiveVehicleReID,liu2017provid} also have large intraclass variation, the number of ids is so small that VeRi-776 is not as complex as Veri-Wild.

As shown in Fig.~\ref{fig:data_samples}, different images of one person have different backgrounds, viewpoints, illuminations, etc. And these elements can also cause the large intraclass distance and small interclass distance in Person ReID tasks. Since the Person ReID tasks have the same challenge as the Vehicle ReID tasks have, we also experiment on three Person ReID databases, Market-1501~\cite{zheng2015scalable}, DukeMCMT-reID~\cite{ristani2016performance}, and CUHK03~\cite{li2014deepreid}, to show that the impressive performance of DSAM loss in ReID tasks.

\subsection{Databases addition and Experiment setting}
As Table.~\ref{table:databases} shows, the Vehicle ReID database, Veri-Wild, has three test sets with different numbers of query id and different gallery id. And detailed statistics of the three widely used Person ReID datasets are provided in Table.~\ref{table:databases}.

As for Person ReID tasks, we also use Cumulative matching characteristics (CMC) Rank-1 accuracy and mean-average-precision (mAP) to evaluate the performance of DSAM loss. In all the Person ReID experiments, the input images resized to 256$\times$128 pixels before sending them to the network. We use the ImageNet pre-trained OSNet~\cite{zhou2019omni} as the basic network architecture. For DSAM loss, we set margin to $0.9$, $\gamma$ to $0.8$, and $\lambda$ to $0.05$ by cross-validation. We use the batch-hard strategy as the triplet loss in our experiments according to ~\cite{kuma2019vehicle}. Each batch contains $32$ different IDs, and each ID contains $8$ different images. During training, we use the SGD optimizer by setting the momentum to $0.9$ and the weight decay to $0.0005$. The learning rate starts from $0.01$ and is divided by $10$ every $10$ epochs. When the model converges, the learning rate is $0.00001$. During testing, we use the output before the loss function as the feature representations and use the cosine distance to calculate the similarity between the query and gallery images.

%-------------------------------------------------------------------------

\section{More comparisons}
\subsection{Comparisons on Veri-Wild}
Table.~\ref{table:loss_wild} shows that the mAP score of ``S+D" loss is much higher than the mAP score of baseline loss and Arcface loss. Moreover, in such Vehicle ReID databases with huge intra-class variation, compared to the softmax, the Angular Margin (Arcface) Loss may not improve the performance. However, our DSAM loss still boosts the mAP by almost 18\% in the small test datasets. The triplet loss is useful, but our DSAM loss outperforms it by a large margin. The triplet contributes little on a large scale test set where mAP is only increased by 0.58\%. Meanwhile, our DSAM loss can improve the baseline by 11.13\% on a large scale test set.

In the Table.~\ref{table:sota_wild}, we compare our result with the recent state-of-the-arts result on Veri-Wild and the Resnet50 trained by ``S+D" performance well.

\subsection{Comparisons on DukeMTMC-reID and Market-1501}
As Table.~\ref{table:sota_duke} shows, the DSAM loss outperforms the triplet loss a large margin. We use OSNet~\cite{zhou2019omni} as our backbone network. The comparison between the results of the OSNet and the results of the OSNet trained by DSAM loss and SoftMax shows that the DSAM loss improve the performance of OSNet by 5.1\%/2.6\% on mAP/Rank-1 on DukeMTMC-reID and by 2.78\%/0.5\% on mAP/Rank-1 on Market-1501. As for trplet loss, our DSAM loss beats the triplet loss by 2.0\%/1.1\% on mAP/Rank-1 on DukeMTMC-reID and by 1.6\%/0.7\% on mAP/Rank-1 on Market-1501. With re-ranking, the results of OSNet trained by DSAM loss outperform many much more complex backbone networks and achieve state-of-the-art performance. The impressive performance surpasses the second place, ISP~\cite{zhu2020identity}, with a considerable margin on both DukeMTMC-reID and Market-1501.

%\subsection{Comparisons on Market-1501}

\subsection{Comparisons on CUHK03}
 As shown in Table.~\ref{table:sota_cuhk}, the OSNet trained by DSAM loss surpasses the original OSNet by 6.6\%/5.0\% on mAP/Rank-1 in CUHK03-Detected dataset. As for mAP, the DSAM loss beats triplet loss by 4.3\% on CUHK03-Labeled and 5.3\% on CUHK03-Detected. With re-ranking, the results of OSNet trained by DSAM loss achieve state-of-the-art performance on both CUHK03-Labeled and CUHK03-Detected.
 
The results show in Table.~\ref{table:sota_duke} and Table.~\ref{table:sota_cuhk} can prove that the DSAM loss can also improve the performance of Person ReID tasks and surpass the triplet loss with a great margin. In general, the DSAM loss is suitable for all ReID tasks and can achieve impressive performance.

%\clearpage

\bibliography{AAAI21ReID}

\end{document}